\documentclass[10pt]{article} 
\usepackage[preprint]{tmlr}

\usepackage{amsmath,amsfonts,bm}

\def\eqref#1{equation~\ref{#1}}

\def\1{\bm{1}}

\DeclareMathAlphabet{\mathsfit}{\encodingdefault}{\sfdefault}{m}{sl}
\SetMathAlphabet{\mathsfit}{bold}{\encodingdefault}{\sfdefault}{bx}{n}

\usepackage[backref=page,hidelinks,colorlinks=true,citecolor=blue]{hyperref}
\usepackage{url}

\usepackage{xspace,mfirstuc,tabulary}
\usepackage[utf8]{inputenc} 
\usepackage[T1]{fontenc}    
\usepackage{hyperref}       
\usepackage{url}            
\usepackage{booktabs}       
\usepackage{amsfonts}       
\usepackage{nicefrac}       
\usepackage{microtype}      
\usepackage{xcolor}         
\usepackage{enumitem}

\usepackage{multicol}
\usepackage{multirow}
\usepackage{makecell}

\usepackage{latexsym}
\usepackage{graphicx}
\usepackage{stfloats}
\usepackage{tablefootnote}

\title{Few-shot Learning with \\ Retrieval Augmented Language Models}

\author{\name Gautier Izacard\thanks{equal contribution}$^{*, \diamondsuit, \clubsuit, \heartsuit}$ \email gizacard@fb.com \\
  \name Patrick Lewis$^{*, \diamondsuit}$ \email plewis@fb.com \\
  \name Maria Lomeli$^\diamondsuit$ \email marialomeli@fb.com \\
  \name Lucas Hosseini$^\diamondsuit$ \email hoss@fb.com \\
  \name Fabio Petroni$^\diamondsuit$ \email fabiopetroni@fb.com \\
  \name Timo Schick$^\diamondsuit$ \email schick@fb.com \\
  \name Jane Dwivedi-Yu$^\diamondsuit$ \email janeyu@fb.com \\
  \name Armand Joulin$^\diamondsuit$ \email ajoulin@fb.com \\
  \name Sebastian Riedel$^{\diamondsuit, \spadesuit}$ \email sriedel@fb.com \\
  \name Edouard Grave$^\diamondsuit$ \email egrave@fb.com \\
  \addr $^\diamondsuit$ Meta AI Research,
  $^\clubsuit$ ENS, PSL University,
  $^\heartsuit$ Inria,
  $^\spadesuit$ University College London
}

\newcommand{\Atlas}{\textsc{Atlas}}

\newcounter{rqcounter}
\newcommand{\RQ}{\stepcounter{rqcounter} \textbf{(RQ \arabic{rqcounter})}}

\def\month{MM}  
\def\year{YYYY} 
\def\openreview{\url{https://openreview.net/forum?id=XXXX}} 
\begin{document}

\title{\Atlas{}: Few-shot Learning with \\ Retrieval Augmented Language Models}

\maketitle

\begin{abstract}%
  Large language models have shown impressive few-shot results on a wide range of tasks.
  However, when knowledge is key for such results, as is the case for tasks such as question answering and fact checking, massive parameter counts to store knowledge seem to be needed.
  Retrieval augmented models are known to excel at knowledge intensive tasks without the need for as many parameters, but it is unclear whether they work in few-shot settings.
  In this work we present \Atlas{}, a carefully designed and pre-trained retrieval augmented language model able to learn knowledge intensive tasks with very few training examples.
  We perform evaluations on a wide range of tasks, including MMLU, KILT and NaturalQuestions, and study the impact of the content of the document index, showing that it can easily be updated.
  Notably, \Atlas{} reaches over 42\% accuracy on Natural Questions using only 64 examples, outperforming a 540B parameters model by 3\% despite having 50x fewer parameters.
\end{abstract}

\section{Introduction}
Large language models~(LLMs) are impressive few-shot learners~\citep{brown2020gpt3,rae2021goepher,hoffmann2022chinchilla,chowdhery2022palm}.
They are able to learn new tasks with very few examples or even from instructions alone.
For this generalisation ability to emerge, the key ingredients are scaling both the parameter count of the model, and the size of the training data.
Large language models owe this improvement to both a larger computational budget, enabling more complex reasoning, and the ability to memorize more information related to downstream tasks from the larger training data.
While it is intuitive to assume that increased reasoning abilities lead to better generalisation, and hence few-shot learning, the same is not true for in-parameter memorisation.
Specifically, it is unclear to what extent effective few-shot learning requires vast knowledge in the parameters of the model.

In this paper, we investigate whether few-shot learning requires models to store a large amount of information in their parameters, and if memorisation can be decoupled from generalisation.
To do so, we leverage the fact that memory can be outsourced and replaced by an external non-parametric knowledge source by employing a \emph{retrieval-augmented} architecture. 
These models employ a non-parametric memory, e.g. a neural retriever  over a large, external, potentially non-static knowledge source to enhance a parametric language model.
In addition to their memorisation abilities, such architectures are attractive due to a number of other established advantages in terms of adaptability, interpretability and efficiency~\citep[][inter alia]{guu2020realm, lewis2020retrieval, yogatama-etal-2021-adaptive,borgeaud2021retro}.
However, retrieval-augmented models have yet to demonstrate compelling few-shot learning capabilities.
In this work we address this gap, and present \Atlas{}, a retrieval-augmented language model capable of strong few-shot learning, despite having lower parameter counts than other powerful recent few-shot learners.

\Atlas{} retrieves relevant documents based on the current context by using a general-purpose dense retriever using a dual-encoder architecture, based on the Contriever~\citep{izacard2022unsupervised}.
The retrieved documents are processed, along with the current context, by a sequence-to-sequence model using the Fusion-in-Decoder architecture~\citep{izacard2020leveraging} that generates the corresponding output. 
We study the impact of different techniques to train \Atlas{} on its few-shot performance on a range of downstream tasks, including question answering and fact checking.
We find that jointly pre-training the components is crucial for few-shot performance, and we carefully evaluate a number of existing and novel pre-training tasks and schemes for this purpose.
\Atlas{} achieves strong downstream performance in both few-shot and resource-rich settings.
For example, with only 11B parameters, \Atlas{} achieves an accuracy of 42.4\% on NaturalQuestions using 64 training examples (45.1\% with a Wikipedia-only index), outperforming PaLM~\citep{chowdhery2022palm}, a 540B parameter model by almost 3 points, and 64.0\% in a full-dataset setting with a Wikipedia index, establishing a new state of the art by 8 points.

\begin{figure}[t]
\centering
\includegraphics[width=.9\textwidth]{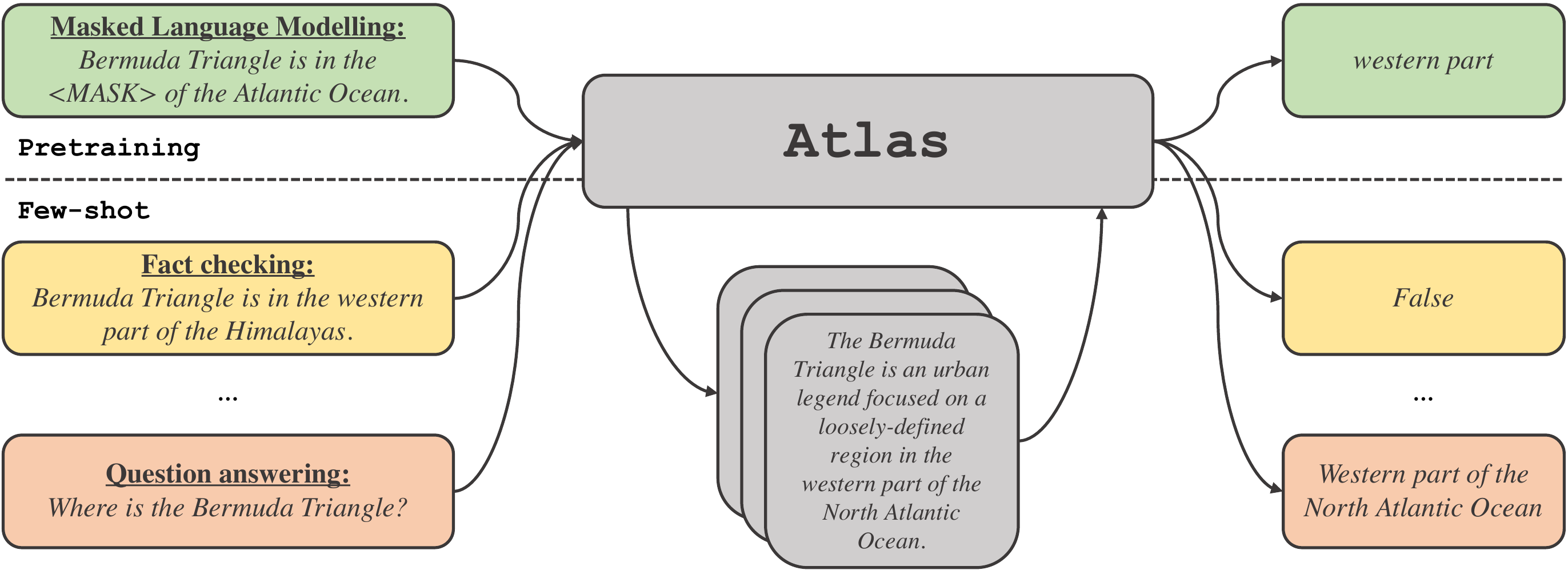}
\caption{We introduce \Atlas, a retrieval-augmented language model that exhibits strong few-shot performance on knowledge tasks, and uses retrieval during both pre-training and fine-tuning.}
\end{figure}

In summary we make the following contributions:
\begin{itemize}
\itemsep0.1em 
\item A thorough study on how to design and train retrieval-augmented language models, with a focus on downstream few-shot learning and sample efficiency.
\item The findings of this study lead to a retrieval-augmented language model, called \Atlas{}, that exhibits few-shot abilities that emerge at lower scale than standard LLM.
\item We provide an exploration of fine-tuning strategies to efficiently adapt both the retriever and the language model to the task at hand.
\item Thorough downstream experiments in few-shot settings, demonstrating state-of-the-art results on few-shot NaturalQuestions (+2.8\%), TriviaQA (+3.3\%), FEVER (+5.1\%), and results on par or stronger than models with 15$\times{}$ more parameters on MMLU.
\item Experiments investigating full-dataset finetuning, setting new state-of-the-art results in NaturalQuestions (+8.1\%), TriviaQA (+9.3\%) and 5 KILT Tasks. 
\item Experiments demonstrating the updatability and interpretability characteristics of \Atlas.
\item Experiments demonstrating that a compressed index using product quantisation achieves comparable performance as an uncompressed index while resulting in a 5x memory reduction.
\end{itemize}

Our code, pretrained \Atlas{} checkpoints, and various supporting data are available at \url{https://github.com/facebookresearch/atlas}

\section{Method}
Our approach follows the \emph{text-to-text} framework~\citep{raffel2019exploring}.
This means that all the tasks are framed as follows: the system gets a \emph{text query} as input, and generates a \emph{text output}.
For example, in the case of question answering, the query corresponds to the question and the model needs to generate the answer.
In the case of classification tasks, the query corresponds to the textual input, and the model generates the lexicalized class label, i.e. the word corresponding to the label.
We give more examples of downstream tasks, from the KILT benchmark in Figure~\ref{fig:kilt_examples}.
As many natural language processing tasks require \emph{knowledge}, our goal is to enhance standard text-to-text models with retrieval, which, as we hypothesise in the introduction, may be crucial to endow models with few-shot capabilities. 

\subsection{Architecture}
Our model is based on two sub-models: the \emph{retriever} and the \emph{language model}.
When performing a task, from question answering to generating Wikipedia articles, our model starts by retrieving the top-k relevant documents from a \emph{large corpus} of text with the retriever.
Then, these documents are fed to the language model, along with the query, which in turns generates the output.
Both the retriever and the language model are based on pre-trained transformer networks, which we describe in more detail below.

\paragraph{Retriever.}
Our retriever module is based on the Contriever~\citep{izacard2022unsupervised}, an information retrieval technique based on continuous dense embeddings.
The Contriever uses a dual-encoder architecture, where the query and documents are embedded independently by a transformer encoder~\citep{huang2013learning,karpukhin2020dense}.
Average pooling is applied over the outputs of the last layer to obtain one vector representation per query or document.
A similarity score between the query and each document is then obtained by computing the dot product between their corresponding embeddings.
The Contriever model is pre-trained using the MoCo contrastive loss~\citep{he2020momentum}, and uses unsupervised data only.
As shown in the following section, an advantage of dense retrievers is that both query and document encoders can be trained without document annotation, using standard techniques such as gradient descent and distillation.

\begin{figure}[t]
  \centering
  \begin{tabular}{p{3cm} p{8cm} p{4cm}}
    \toprule
    \textbf{Task} & \textbf{Query} & \textbf{Output} \\
    \midrule
    Fact Checking & \textit{Bermuda Triangle is in the western part of the Himalayas.} & \textit{False} \\
    \midrule
    Question Answering & \textit{who is playing the halftime show at super bowl 2016} & \textit{Coldplay} \\
    \midrule
    Entity Linking & \textit{NTFS-3G is an open source <E>cross-platform</E> implementation of the Microsoft Windows NTFS file system with read-write support.} & \textit{Cross-platform software} \\
    \bottomrule
  \end{tabular}
  \caption{Examples of query and output pairs for different tasks from KILT.}
  \label{fig:kilt_examples}
\end{figure}

\paragraph{Language model.}
For the language model, we rely on the T5 sequence-to-sequence architecture~\citep{raffel2019exploring}.
We rely on the Fusion-in-Decoder modification of sequence-to-sequence models, and process each document independently in the encoder~\citep{izacard2020leveraging}.
We then concatenate the outputs of the encoder corresponding to the different documents, and perform cross-attention over this single sequence in the decoder.
Following \citet{izacard2020leveraging}, we concatenate the query to each document in the encoder.
Another way to process the retrieved documents in the language model would be to concatenate the query and all the documents, and to use this long sequence as input of the model.
Unfortunately, this approach does not scale with the number of documents, since the self-attention in the encoder results in a quadratic complexity with respect to the number of documents.

\subsection{Training objectives for the retriever}
\label{sec:retriever_loss}
In this section, we discuss four different loss functions to train the retriever jointly with the language model.
We consider loss functions that leverage the language model to provide \emph{supervisory signal} to train the retriever.
In other words, if the language model finds a document useful when generating the output, the retriever objective should encourage the retriever to rank said document higher.
This allows us to train models using only query and output pairs from the task of interest, without relying on document annotations.
For example, in the case of fact checking, a model only requires pairs of claims and corresponding verdicts but no documents containing the evidence to back up the verdict.
In practice, we can apply this approach on any task, including self-supervised pre-training.
As shown in the experimental section, pre-training is critical for obtaining models that exhibit few-shot learning abilities.

\paragraph{Attention Distillation (ADist).}
The first loss that we consider is based on the attention scores of the language model, and is heavily inspired by~\citet{izacard2021distilling}.
The main idea is that the cross-attention scores between the input documents and the output, can be used as a proxy of the importance of each input document when generating the output.
In particular, \citet{izacard2021distilling} showed that these scores can be aggregated across attention heads, layers and tokens for a given document to obtain a single score for each document.
Then, these scores can be distilled into the retriever by minimizing the KL-divergence with the probability distribution $p_{\textsc{retr}}$ over the top-K documents $\{\mathbf{d}_k\}_{1, ..., K}$ obtained from the retriever:
\begin{equation}
  p_{\textsc{retr}}\left(\mathbf{d} \ | \ \mathbf{q}\right) = \frac{\exp(s(\mathbf{d}, \mathbf{q}) / \theta)}{\sum_{k=1}^K \exp(s(\mathbf{d}_k, \mathbf{q}) / \theta)},
  \label{eqn:prob_retr}
\end{equation}
where $s$ is the dot-product between the query and documents vectors and $\theta$ is a temperature hyper-parameter.

In the original paper, it was proposed to use the pre-softmax scores from the decoder cross-attentions, and average across heads, layers and tokens.
Here, we propose an alternative which gives slightly stronger results, which relies on the following observation.
In the attention mechanism, as defined by
$$
\mathbf{y} = \sum_{n=1}^N  \alpha_n \mathbf{v}_n,
$$
the contribution to the output $\mathbf{y}$ of a particular token $n$ cannot be evaluated from the attention score $\alpha_n$ alone,
but should also take the norm of the value $\mathbf{v}_n$ into account.
Hence, we use the quantity $\alpha_n \| \mathbf{v}_n \|_2$ as the measure of relevance for token $n$.
Following \citet{izacard2021distilling}, we average these scores over all attention heads, layers, and tokens to obtain a score for each document.
We apply the \textsc{Softmax} operator over the resulting scores, to obtain a distribution $p_{\textsc{attn}}(\mathbf{d}_k)$ over the top-K retrieved documents.
We then minimize the KL-divergence between  $p_{\textsc{attn}}(\mathbf{d}_k)$, and the distribution $p_{\textsc{retr}}$ from the retriever defined in Equation~\ref{eqn:prob_retr}:
$$
\textsc{KL}(p_{\textsc{attn}} \ \| \ p_{\textsc{retr}}) = \sum_{k=1}^K p_{\textsc{attn}} (\mathbf{d}_k) \log \left( \frac{p_{\textsc{attn}} (\mathbf{d}_k)}{p_{\textsc{retr}} (\mathbf{d}_k)} \right).
$$
Here, this loss is only used to optimize the parameters of the retriever, and not the language model.
When using recent deep learning frameworks, this is achieved by applying a \textsc{StopGradient} operator on $p_{\textsc{attn}}$.

\paragraph{End-to-end training of Multi-Document Reader and Retriever (EMDR$^2$).}
Next, we consider the method introduced by~\citet{emdr}, which is inspired by the expectation-maximization algorithm, treating retrieved documents as latent variables.
Given a query $\mathbf{q}$, the corresponding output $\mathbf{a}$ and the set $\mathcal{D}_K$ of top-K retrieved documents with the current retriever, the EMDR$^2$ loss to train the retriever is
$$ \log \left[ \sum_{k=1}^K p_{\textsc{lm}}(\mathbf{a} \ | \ \mathbf{q}, \mathbf{d}_k) p_{\textsc{retr}}(\mathbf{d}_k \ | \ \mathbf{q}) \right], $$
where $p_{\textsc{retr}}$ is again the probability over the top-K documents obtained with the retriever, as defined by Equation~\ref{eqn:prob_retr}.
Again, only the parameters of the retriever are updated
by applying a \textsc{StopGradient} operator around $ p_{\textsc{lm}}$.
One should note that the probability distribution over documents that maximizes this loss function is an indicator of the document corresponding to the highest probability of the output according to the language model.
Finally, in practice, the EMDR$^2$ loss function is applied at the token level, and not at the sequence level.

\paragraph{Perplexity Distillation (PDist).}
Third, we discuss a simpler loss function which is loosely inspired by the objectives from the attention distillation and EMDR$^2$ methods~\citep{izacard2021distilling,emdr}.
More precisely, we want to train the retriever to predict how much each document would improve the language model perplexity of the output, given the query.
To this end, we minimize the KL-divergence between the documents distribution of the retriever (Eqn.~\ref{eqn:prob_retr}), and the documents posterior distribution according to the language model, using a uniform prior:
$$
p_k \propto p_{LM} (\mathbf{a} \ | \ \mathbf{d}_k, \mathbf{q}).
$$
Using the \textsc{Softmax} operator, we have that
$$
p_k = \frac{\exp(\log p_{LM} (\mathbf{a} \ | \ \mathbf{d}_k, \mathbf{q}))}{\sum_{i=1}^K \exp ( \log p_{LM} (\mathbf{a} \ | \ \mathbf{d}_i, \mathbf{q}))}.
$$

\paragraph{Leave-one-out Perplexity Distillation (LOOP).}
Finally, we propose an objective based on how  much \emph{worse} the prediction of the language model gets, when \emph{removing} one of the top-k retrieved documents.
To do so, we compute the log probability of the output for each subset of k-1 documents, and use the negative value as relevance score for each document.
Following the previous loss function, we use the softmax operator to obtain a probability distribution over documents:
$$
p_{\textsc{loop}}(\mathbf{d}_k) = \frac{\exp(- \log p_{LM} (\mathbf{a} \ | \ \mathcal{D}_K \setminus \{ \mathbf{d}_k \}, \mathbf{q}))}{\sum_{i=1}^K \exp (- \log p_{LM} (\mathbf{a} \ | \ \mathcal{D}_K \setminus \{ \mathbf{d}_i \}, \mathbf{q}))}.
$$
As before, we then minimize the KL-divergence between this distribution, and the one obtained with retriever.
This loss is more expensive to compute than PDist and EMDR, but, like ADist, employs the language model more closely to the way it is trained i.e. the LM is trained to be conditioned on a set of $K$ documents. 
For LOOP, the language model is conditioned on $(K-1)$ documents, rather than a single document as in EMDR$^2$ and PDist. 

For all losses, we can also use a temperature hyper-parameter when computing the target or retriever distributions to control the distribution's peakiness of, which might be important for some tasks or losses. 
Indeed, for PDist and LOOP, the perplexity of the output may not vary much when conditioning on different documents, especially in the case of long outputs.

\subsection{Pretext tasks}
\label{section:pretext_tasks}
In this section, we describe pretext tasks that can be used to jointly pre-train the retriever and the language model using only unsupervised data.

\paragraph{Prefix language modeling.}
First, we consider a standard language modeling task as potential pre-training objective.
To cast language modeling in the text-to-text framework, we consider a chunk of $N$ words, and split this chunk in two sub-sequences of equal length $N/2$.
Then, the first sub-sequence is used as the query, and the second corresponds to the output.
We thus retrieve relevant documents by using the first sub-sequence of $N/2$ tokens, to generate the output.

\paragraph{Masked language modeling.}
Second, we consider masked language modeling, as formulated by \citet{raffel2019exploring}.
Again, starting from a chunk of $N$ words, we sample $k$ spans of average length 3 tokens, leading to a masking ratio of $15\%$.
We then replace each span by a different special token.
The model is then trained to generate the masked spans, each span beginning with the special sentinel mask token that was inserted in the input sequence.
We retrieve documents using the masked query, but replace the special mask tokens with a mask token supported by the retriever vocabulary.

\paragraph{Title to section generation.}
Finally, we consider a more abstractive generation task, generating sections from Wikipedia articles, given the article and section title.
Here, the query corresponds to the title of the article, together with the title of the section, and the output corresponds to the text of the section.
We exclude sections ``See also'', ``References'', ``Further reading'' and ``External links''.

\subsection{Efficient retriever fine-tuning}
\label{sec:efficient_finetuning}

Retrieval is facilitated by using a document \emph{index}, which is a pre-computed collection of the document embeddings for all the documents in the retrieval corpus.
When jointly training the retriever and language model, the index
needs to be updated regularly, otherwise, the embeddings of the documents stored in the index become stale relative to the updated retriever.
This means that we need to recompute the embeddings for the full collection of documents regularly during training to keep the index fresh, which can be computationally expensive for large indices.
This is particularly true at \emph{fine-tuning} time, where the number of training examples could be small relative to the number of documents in the index.
Training the retriever could thus add an important computational overhead compared to standard language model finetuning.
In this section, we analyse strategies that might make this process more efficient, alleviating the need to re-compute the embeddings of all the documents too often.

\paragraph{Full index update.}
Let us start by analysing the overhead due to updating the index, compared to using a fixed retriever.
To compare the computation time of different models, we will make the following assumption:
the time required to perform a forward pass on a document with a model of $P$ parameters is $O(P)$.
While this computation model may seem naive, the main assumption is that document sizes are constant.\footnote{See \citet{hoffmann2022chinchilla} for more details about the computation of the FLOPS corresponding to the forward and backward passes of transformer networks.}
Since we split long documents into passages with similar number of words, and use padding when processing documents of different sizes, this assumption is reasonable in practice.
Let $K$ be the number of documents that are retrieved and processed by the language model, $P_{\textsc{lm}}$ be the number of parameters of the language model and $B$ the batch size.
Each training step has a complexity of $4 \times B \times K \times P_{\textsc{lm}}$.\footnote{There is a factor 4 to account for the backward pass and activation checkpointing.}

Next, let $N$ be the number of documents in the index, and $P_{\textsc{retr}}$ be the number of parameters of the retriever.
Then, re-computing the full index has a complexity of $N \times P_{\textsc{retr}}$.
If we refresh the index every $R$ training steps, we obtain the following overhead:
$$
\frac{N \times P_{\textsc{retr}}}{4 \times B \times K \times P_{\textsc{lm}} \times R}.
$$
If we use the BERT-base architecture for our retriever and T5-XL for our language model, we get $\frac{P_{\textsc{retr}}}{P_{\textsc{lm}}} \approx \frac{1}{25}$, lading to the overhead:
$$
\frac{N}{100 \times B \times K \times R}.
$$
If we use an index containing $37M$ documents (the size of our Wikipedia index), train with a batch size of $64$ with $20$ retrieved documents and refresh the index every 1000 steps, this results in an overhead of $\sim 30\%$. 

\paragraph{Re-ranking.}
A second strategy is to retrieve a larger number of documents $L$ with the retriever, and to re-embed and rerank these documents with the up-to-date retriever, and pass the resulting top-$K$ to the language model.
In that case, the overhead of reranking the top-$L$ documents is equal to $B \times L \times P_{\textsc{retr}}$.
Since we perform this operation at every time step, the overhead is equal to
$$\frac{L \times P_{\textsc{retr}}}{4 \times K \times P_{\textsc{lm}}}.$$
Using the same assumption as before, we finally get that the overhead is of the order of  $\frac{L}{100 \times K}$.
If we re-rank 10x more documents than what the language model processes (i.e., $L = 10 \times K$), we get an overhead of $10\%$.
However, note that if many updates are performed on the retriever, the index might still need to be fully updated, as the true top-k documents may not be retrieved in the top-L results from the stale index.
In practice, it is possible to track the positions of the top-K re-ranked documents in the top-L, and estimate when the index needs to be updated.

\paragraph{Query-side fine-tuning.}
Finally, the last strategy is to decouple the encoding of the queries and documents.
In this case, we fix the parameters corresponding to the document encoder, and only train the parameters corresponding to the query encoder.
Thus, the embeddings of documents are fixed, and we do not need to refresh the index, and thus there is no computational overhead.
As we will see in practice, the impact of fixing the documents encoder varies greatly for different tasks when a large training dataset is available.
For most of the few-shot settings that we consider, query-side finetuning does not have large performance impact, and sometimes even slightly improves performance.

\section{Related work}

\subsection{Retrieval in natural language processing}
\paragraph{Retrieval for knowledge intensive tasks.}
Previous work has shown that retrieval improves performance across a variety of tasks such as question answering~\citep{voorhees1999trec,chen2017reading, kwiatkowski2019natural}, fact checking~\citep{thorne2018fever}, dialogue~\citep{dinan2019wizardofwikipedia} or citation recommendation~\citep{petroni2022improving}.
Historically, this information retrieval step was implemented using term-matching methods, such as TF-IDF or BM25~\citep{jones1972statistical,robertson1995okapi}.
For open-domain question answering~\citep{voorhees1999trec}, documents are often retrieved from Wikipedia~\citep{chen2017reading}.
Recently, dense retrievers based on neural networks have become popular.
These usually follow a dual-encoder architecture~\citep{yih_learning_2011,huang2013learning,shen2014learning}, where queries and passages are encoded independently as vectors, and relevance is computed using the inner product or Euclidean distance.
Popular supervised retrievers include DPR~\citep{karpukhin2020dense}, which is trained to discriminate the relevant passage among negative passages, and extensions such as ANCE~\citep{xiong2020approximate} which improved the hard negatives mining process.
We refer the reader to \citet{yates_pretrained_2021} for a survey of dense retrieval techniques.

After retrieval, the relevant documents are processed to produce the final output.
In open-domain QA, models can extract a span of text from retrieved documents as the answer~\citep{chen2017reading, clark2017simple, wang2019multi, karpukhin2020dense}, a method inspired by reading comprehension~\citep{richardson_mctest:_2013,rajpurkar2016squad}.
Recently, generating the answer as free-form text, using a seq2seq model conditioned on retrieved documents have become prevalent~\citep{lewis2020retrieval,izacard2020leveraging,min2020ambigqa}.
These architectures have also been shown to reduce hallucination in dialogue agents~\citep{shuster2021retrieval}.

\paragraph{Retriever training.}
The need for expensive query-document annotations for training the retriever can be bypassed, by leveraging signals from the language model, or using unsupervised learning.
REALM~\citep{guu2020realm} and RAG~\citep{lewis2020retrieval} jointly train the retriever and language model by modelling documents as latent variable, and minimizing the objective with gradient descent. REALM pre-trains end-to-end with an MLM approach but uses an extractive BERT-style model~\citep{devlin2018bert}. 
\citet{guu2020realm} also explore a query-side finetuning at finetuning time to avoid index refreshes, which is also explored in the context of phrase-based retrieval by \citet {lee-etal-2021-learning-dense}. 
\citet{izacard2020leveraging} proposed to use cross-attention scores as supervision with knowledge distillation.
\citet{emdr} perform joint training of the reader and the retriever by leveraging the perplexity of the output generated by the reader.
\citet{emdr} and \citet{yono} both employ salient span masking to pre-train retrievers, leveraging the perplexity and attention scores from the language model.
The \emph{inverse cloze task} was proposed by \citet{lee2019latent} to pre-train dense retrievers in an unsupervised way.
\citet{paranjape2021hindsight} propose a method to train retrieval-augmented generators using a second ``informed'' retriever with access to the output, which the test-time retriever can be distilled from, and \citet{Hofsttter2022MultiTaskRT} recently proposed a training set filtering/weighting approach to train stronger retrieval-augmented generators. 
\citet{izacard2022unsupervised} explored different contrastive learning methods to train retrievers, while \citet{ram-etal-2022-learning} used recurring spans within a document to create pseudo-positive query-document pairs.

\paragraph{Retrieval-augmented language models.}
Continuous cache models~\citep{graveImproving2016} defines a probability distribution over recent tokens, by computing the similarity between previous and current representations of tokens.
This distribution is then interpolated with the distribution of the language model, to improve predictions.
Later, the amount of tokens used to compute this distribution was extended to a much larger memory by leveraging approximate nearest neighbors search~\citep{grave2017unbounded}.
The related kNN-LM model~\citep{khandelwalGeneralization2020} replaced LSTMs by transformer networks, and scaled the memory to billions of tokens, leading to strong performance improvements.
More recently, RETRO~\citep{borgeaud2021retro} extended these by scaling the retrieval memory to trillions of tokens, and changing the model architecture to take retrieved documents as input.

\paragraph{Retrieval-Augmentation with Search Engines.}
Recently, different works have proposed to train large language models to interact with a search engine, by generating text queries, and using the retrieved documents as additional context~\citep{Nakano2021WebGPTBQ,Thoppilan2022LaMDALM,Shuster2022LanguageMT}.
In the context of few-shot question answering, \citet{Lazaridou2022InternetaugmentedLM} used the question to perform a search query, and retrieved documents are added to the prompt of a large language model performing in-context learning.

\subsection{Few-shot learning}
Few-shot learning, the task of learning from very few examples, has been studied for decades \citep{thrun_learning,fink_object,matchingnetworks}, but has recently seen an explosion of interest in NLP with the arrival of large pre-trained models, which exhibit emergent few-shot learning abilities~\citep{wei2022emergent}.

\paragraph{In-context Learning with large Language models.}
Providing language models with natural language descriptions of tasks, as proposed by ~\citet{radford2019language} has led to significant developments in few-shot learning.
GPT-3~\citep{brown2020gpt3} demonstrated the ability of large language models to perform few-shot predictions, where the model is given a description of the task in natural language with few examples.
Scaling model size, data and compute is crucial to enable this learning ability, leading to the further development of large models~\citep{lieber2021j1, rae2021goepher, smith2022megatron, chowdhery2022palm, smith2022megatron}.
\citet{hoffmann2022chinchilla} revisited the scaling law from~\citet{kaplan2020scaling}, suggesting that training on more data with a smaller model may be more effective, resulting in Chinchilla, a 70B parameter model with improved parameter efficiency.

\paragraph{Few-shot finetuning and prompt-based learning.}
The above models perform few-shot learning with in-context instructions without training the parameters of the language model.
Few-shot learning can also be accomplished by combining textual templates (``prompts'') and various forms of model finetuning, either fully updating a model's parameters, e.g. for classification~\citep{schick2021classification,Schick2021ItsNJ,gao-etal-2021-making,Tam2021ImprovingAS} or generation \citep{schick2021generation}.
Prompts themselves can be optimized, for example by search \citep{jiang-etal-2020-know, shin-etal-2020-autoprompt} or by only updating parts of the model~\citep{RobertLLogan2021CuttingDO}, or learning ``soft-prompts''~\citep{prompt_tuning,li-liang-2021-prefix}.
Due to its simplicity, in this work we either employ simple prompts or simply feed in inputs without preprocessing, and perform full-model finetuning, a method similar to \citet{le-scao-rush-2021-many}.

\section{Experiments}
In this section, we report empirical evaluations of our language models on few-shot learning.
We start by introducing our experimental setup, describing our evaluation benchmarks in section~\ref{sec:benchmarks}, and giving the training details of our models in section~\ref{sec:training}.
Then, we perform an ablation study to compare the different technical choices leading to our main model.
We finally evaluate this model, called \Atlas, on different natural language understanding tasks in few-shot and full dataset settings.

\subsection{Benchmarks}
\label{sec:benchmarks}
To evaluate our retrieval-augmented language models we consider the following benchmarks, which include different tasks.

\paragraph{Knowledge-Intensive Language Tasks (KILT).} First, we use the KILT evaluation suite~\citep{petroni2020context}, containing 11 datasets corresponding to 5 tasks: fact checking, question answering, dialog generation, entity linking and slot-filling.
These different tasks require knowledge about the world to be solved, which can be found in Wikipedia.
We evaluate our model on the following tasks and datasets included in KILT: question answering: NaturalQuestions~\citep{kwiatkowski2019natural}, TriviaQA~\citep{JoshiTriviaQA2017} and HotpotQA~\citep{yang2018hotpotqa}; slot filling: Zero~Shot~RE~\citep{levy2017zero} and T-REx~\citep{elsahar2018trex}; entity linking: AIDA CoNLL-YAGO~\citep{hoffart2011robust}; dialogue: Wizard of Wikipedia~\citep{dinan2019wizardofwikipedia}; and fact checking: FEVER~\citep{thorne2018fever}.
The KILT versions of these datasets differ from their original versions, as instances requiring knowledge not present in the August 2019 Wikipedia dump have been removed.

\paragraph{Massively-Multitask Language Understanding (MMLU).} Our second main evaluation benchmark is MMLU~\citep{hendrycks2021mmlu}, which contains 57 multi-choice question answering datasets (referred to as domains), sourced from real examinations designed for humans.
These cover a very broad range of topics, e.g. high school mathematics, professional law, logical fallacies and clinical knowledge and can be broadly categorized in four subsets: humanities, social sciences, STEM and ``other''.
We focus on few-shot learning, and the authors of the benchmarks suggest to use 5 training examples per domain.
Beyond the 5-shot setting, We also consider three additional settings.
The first is a \emph{zero-shot} setting, with no training data at all.
The second, which we call \emph{multi-task few-shot}, is where we train a single model on the 6-shot data from all tasks, hence leading to a training set of 285 examples.
The last, which we call \emph{transfer learning}, leverages additional training examples from other multiple-choice QA tasks provided by the MMLU authors, namely MCTest~\citep{richardson_mctest:_2013}, RACE~\citep{lai-etal-2017-race}, ARC~\citep{Clark2018ThinkYH} and OBQA~\citep{mihaylov-etal-2018-suit} leading to a training set of 95k examples.

\paragraph{Additional benchmarks.} Additionally, we report results on the original open-domain versions of the popular NaturalQuestions~\citep{kwiatkowski2019natural}, and TriviaQA~\citep{JoshiTriviaQA2017} datasets.
We also evaluate our model on the original version of FEVER~\citep{thorne2018fever}, which presents fact checking as a three-way classification problem for textual claims (either ``Supported'': the text is supported by evidence in Wikipedia, ``refuted'': the claim is not consistent with evidence in Wikipedia, or ``not enough info'', where there is insufficient evidence to make a judgement).
We also perform experiments to assess temporal sensitivity of our models. Here, we construct a dataset from TempLAMA~\citep{dhingra-etal-2022-time}, consisting of a set of time-sensitive cloze questions on a range of topics, where the answer changes from 2017 to 2020. 
We assess the accuracy of our models when supplied with a index from 2017 vs 2020 to assess to what degree models faithfully reflect the content of the index supplied to them at test time, and how effective updating the index is as a \emph{continual learning} or model updateability method.

\subsection{Technical details}
\label{sec:training}

We now describe the procedure for pre-training and fine-tuning our models. We focus on the setting used for the ablation studies performed in Section~\ref{sec:loss_tasks} and Section~\ref{sec:finetuning}. 
We give more details about the hyperparameters used for our final model later.

\paragraph{Pre-training.} For the pre-training, we initialize the retriever module using the unsupervised \emph{Contriever} model, which uses the BERT-base architecture.
We initialize the language model with the T5 pre-trained weight.
As the original T5 pre-trained model included supervised data in the training set, we use the version 1.1 models which were trained on unlabeled text only.
Specifically, we initialize from the \texttt{T5-lm-adapt} variants due to their improved stability.

For the ablation studies performed in Section~\ref{sec:loss_tasks} and Section~\ref{sec:finetuning}, we use T5-XL which contains 3B weights.
We pre-train all our models for 10,000 iterations, using AdamW with a batch size of 64 and a learning rate of $10^{-4}$ for the reader and $10^{-5}$ for the retriever with linear decay and 1,000 warmup steps.
We refresh the index every 1,000 steps.
This means that the index is recomputed 10 times during the pre-training, leading to an overhead of around 30\%, compared to training with a fixed retriever.
We set the number of retrieved documents to 20.
We detail the hyperparameters used for the training of our final model at the beginning of Section~\ref{sec:final}.

\paragraph{Fine-tuning.}
When performing a downstream task, either in a few-shot setting or with a large training set, we employ fine-tuning to adapt our models to these tasks.
For the few-shot KILT ablation experiments, we perform a fixed number of fine-tuning iterations, instead of using early-stopping.
More precisely, we decided to use 50 iterations for the 64-shot setting and 200 iterations in the 1024-shot setting.
In both cases, we use a batch size of $32$ examples, a learning rate of $4\times 10^{-5}$ with linear decay and 5 warmup steps for both the reader and the retriever.

\begin{table}[t]
  \centering
  \caption{\textbf{Retriever loss ablation.} 
    We compare different loss functions to pre-train the retriever jointly with the language model.
    We use the prefix MLM task, and the December 2021 Wikipedia dump for both the index and pre-training data.
    Fine-tuning is performed with query-side fine-tuning and the loss used for pre-training. Best result is bold, second highest underlined.
    }
  \vspace{0.5em}
  \label{table:loss_ablation}
  \begin{tabular}{l c cccc cccc}
    \toprule
    && \multicolumn{4}{c}{64-shot} & \multicolumn{4}{c}{1024-shot} \\
    \cmidrule(lr){3-6} \cmidrule(lr){7-10}
    & MLM & NQ & WoW & FEVER & Avg. & NQ & WoW & FEVER & Avg. \\
    \midrule
    Closed-book  & 1.083& 6.5 &14.1 & 59.0 &26.5 & 10.7 & 16.5 & 75.3 & 34.2 \\ 
    No Joint pre-training & -  & 9.0  & 14.1 & 67.0 & 30.0 & 9.9  & 16.6 & 78.3 & 34.9 \\
    Fixed retriever       & 0.823 & 39.9 & 14.3 & 72.4 & 42.2 & 45.3 & \underline{17.9} & 90.0 & \underline{51.1} \\
    ADist                 & \underline{0.780} & 40.9 & 14.4 & 73.8 & 43.0 & \underline{46.2} & 17.2 & \textbf{90.9} & \textbf{51.4} \\
    EMDR$^2$              & 0.783 & \underline{43.3} & \underline{14.6} & 72.1 & 43.3 & 44.9 & \textbf{18.3} & 85.7 & 49.6 \\
    PDist                 & 0.783 & \textbf{45.0} & \textbf{15.0} & \textbf{77.0} & \textbf{45.7} & 44.9 & \underline{17.9} & \underline{90.2} & 51.0 \\
    LOOP                  & \textbf{0.766} & 41.8 & \textbf{15.0} & \underline{74.4} & \underline{43.7} & \textbf{47.1} & \underline{17.9} & 87.5 & 50.8 \\
    \bottomrule
  \end{tabular}
\end{table}

\paragraph{Unlabeled datasets.}
Finally, we discuss the unlabeled text datasets that we use to train our models, which form the retrieval index.
First, we consider the Dec. 20, 2021 Wikipedia dump, for which we keep the lists and infoboxes, which are linearized by adding a semi-colon separator between the entries.
We split articles by section, and split long sections into passages of equal sizes and containing less than 200 words.
This leads to a total of 37M passages, containing 78 words in average.
We also use documents from the 2020-10 common crawl dump, preprocessed with the CCNet pipeline~\citep{wenzek-etal-2020-ccnet}.
We perform additional document filtering, in a similar fashion to Gopher~\citep{rae2021goepher}.
More precisely, we filter documents based on document length, average word length, ratio of alphanumeric characters and number of repeated tokens.
This leads to a total of 350M passages.
The same passages are used for the index and model pre-training.
During pre-training, we ensure the passage we are training on is filtered out from the retrieved documents, to prevent the model from simply retrieving the passage it is de-nosing/generating, and trivially using it to solve the pre-training task.

\subsection{Pre-training loss and tasks}
\label{sec:loss_tasks}
We start our ablation study by comparing different pre-training tasks, and objective functions to jointly train the retriever and the language model.
Our goal here is to answer the following research questions:
\begin{itemize}[leftmargin=5em]
\item[\RQ] Does jointly pre-training the whole model lead to better few-shot performance?
\item[\RQ] What is the best objective function for the retriever, and the best pretext task?
\end{itemize}
We start by comparing the training objectives of the retriever, introduced in Section~\ref{sec:retriever_loss}, by pre-training models using the masked language modelling task.
We evaluate these models on a subset of the 64-shot and 1024-shot KILT benchmark:
NaturalQuestions, FEVER and Wizard of Wikipedia, along with two baselines:  a `closed-book'' (i.e. non-augmented T5) baseline, pre-trained on the same data, and initialized from \texttt{Contriever} and \texttt{T5-lm-adapt}.
We report results in Table~\ref{table:loss_ablation}.
First, we note the poor performance of the closed-book baseline, indicating the importance of augmentation.
Next, we observe that pre-training our model with retrieval is important to obtain good performance on few-shot tasks.
Indeed, all models that include retrieval during pre-training strongly outperform the baseline without joint pre-training.
Next, we compare a model that was pre-trained with a fixed retriever, and models using the various retriever training objectives.
On the MLM validation metric corresponding to the pre-training objective, we observe that jointly training the retriever leads to strong improvements.
This effect tends to be less marked on 64-shot downstream tasks, and almost non-existent for 1024-shot.
We believe that this is evidence that the biggest impact of pre-training is on the language model, which learns to use and aggregate information from the retrieved documents.
Lastly, we do not observe significant systematic differences between the different retriever training objectives.
We thus decide adopt use Perplexity Distillation for subsequent experiments, as it tends to be more stable than EMDR$^2$ or ADist, and more computationally efficient than LOOP.

\begin{table}
  \centering
  \caption{\textbf{Pretext task ablation.} We compare different pretext tasks, used to jointly pre-train our models. Examples are randomly sampled from the training set of the KILT version of the dataset. We report the exact match on NaturalQuestions, the F1 score on Wizard of Wikipedia and the accuracy on FEVER.}
  \vspace{0.5em}
  \label{table:task_ablation}
  \begin{tabular}{l cccc cccc}
    \toprule
    & \multicolumn{4}{c}{64-shot} & \multicolumn{4}{c}{1024-shot} \\
    \cmidrule(lr){2-5} \cmidrule(lr){6-9}
    & NQ & WoW & FEVER & Avg. & NQ & WoW & FEVER & Avg. \\
    \midrule
    Prefix Language Modelling   & 41.0 & 14.5 & 64.9 & 40.1 & \textbf{44.7} & 17.9 & 86.0 & 49.5 \\
    Masked Language Modelling   & \textbf{42.7} & \textbf{14.9} & \textbf{69.7} & \textbf{42.4} & \textbf{44.7} & \textbf{18.3} & \textbf{88.8} & \textbf{50.6} \\
    Title-to-section generation & 41.1 & 15.2 & 66.1 & 40.8 & 45.4 & 17.9 & 84.6 & 49.3 \\
    \bottomrule
  \end{tabular}
\end{table}

Next, we compare the different self-supervised pretext tasks introduced in Section~\ref{section:pretext_tasks}  in Table~\ref{table:task_ablation}.
Here we observe similar results for all three tasks, with a small advantage for masked language modelling.
Thus, in what follows, we adopt masked language modelling for pre-training.

\begin{table}
  \centering
  \caption{\textbf{Index content ablation.} In this table, we report results for models where the content of the index was changed between the pre-training and the fine-tuning.}
  \vspace{0.5em}
  \label{table:index_content_ablation}
  \begin{tabular}{ll cccc cccc}
    \toprule
    & & \multicolumn{4}{c}{64-shot} & \multicolumn{4}{c}{1024-shot} \\
    \cmidrule(lr){3-6} \cmidrule(lr){7-10}
    Index & Training data & NQ & WoW & FEVER & Avg. & NQ & WoW & FEVER & Avg. \\
    \midrule
    Wiki    & Wiki   & \textbf{42.7} & 14.9 & 69.7 & \textbf{42.4} & 44.7 & 18.3 & 88.8 & \textbf{50.6} \\
    Wiki    & CC     & 40.9 & \textbf{15.3} & 67.3 & 41.2 & \textbf{44.8} & \textbf{18.4} & 88.1 & 50.4 \\
    CC      & Wiki   & 32.9 & 14.5 & \textbf{72.1} & 39.8 & 37.8 & 17.1 & 85.8 & 46.9 \\
    CC      & CC     & 38.4 & 14.9 & 70.1 & 41.1 & 42.0 & 17.3 & \textbf{88.9} & 49.4 \\
    \bottomrule
  \end{tabular}
\end{table}

Finally, we consider different combinations of data sources---Wikipedia and common crawl---for the index and training data during pre-training.
In all cases, we use the Wikipedia 2021 dump as the index when performing few-shot fine-tuning.
We report results in Table~\ref{table:index_content_ablation}.
First, we observe that using a Wikipedia-based index leads to better downstream performance.
There could be two explanations for this: first, as we use Wikipedia for the few-shot tasks, the model might be better adapted when trained using the same data.
Another explanation might be that Wikipedia is a higher-quality and denser source of knowledge than common crawl.
Second, when using a common crawl index, we observe that pre-training on Wikipedia data leads to lower performance than using common crawl data.
We believe that the primary reason is that the distribution mismatch between the two domains leads to generally-less relevant retrieved documents.
In turn, this probably means that the pre-training is less efficient, because the language model does not leverage as much information from the documents.
In the following, we thus decide to combine the data from both domains for both the index and the pre-training data.

\subsection{Fine-tuning}
\label{sec:finetuning}
In this section, we perform an ablation study on how to apply our models on downstream tasks, which relies on fine-tuning.
In particular, we want to investigate the following research question:
\begin{itemize}[leftmargin=5em]
\item[\RQ] How to efficiently fine-tune \Atlas~on tasks with limited training data?
\end{itemize}
To answer this question, we compare the different strategies to fine-tune the retriever module, described in Section~\ref{sec:efficient_finetuning}.
We report results in Table~\ref{table:finetuning_ablation}.
First, as for pre-training, we observe that keeping the retriever fixed during fine-tuning leads to a significant performance drops, for both 64- and 1024-shot settings.
Second, the re-ranking strategy (row 2) leads to very similar results to fully updating the index (row 1), while being significantly more efficient.
Lastly, fine-tuning only the query encoder also leads to strong results: in particular, in the 64-shot setup, this is slightly stronger than performing full fine-tuning, which we attribute to there being less opportunity for over-fitting.
On the other hand, on the 1024-shot setting, performing a full fine-tuning leads to stronger results, especially on NaturalQuestions.
In the following, we thus use query-side fine-tuning for experiments with small numbers of examples, and  standard fine-tuning for larger datasets.

\begin{table}
  \centering
  \caption{\textbf{Retriever fine-tuning ablation.} Here, we compare different strategies to fine-tune the retriever in a few-shot setting.}
  \vspace{0.5em}
  \label{table:finetuning_ablation}
  \begin{tabular}{l cccc cccc}
    \toprule
    & \multicolumn{4}{c}{64-shot} & \multicolumn{4}{c}{1024-shot} \\
    \cmidrule(lr){2-5} \cmidrule(lr){6-9}
    & NQ & WoW & FEVER & Avg. & NQ & WoW & FEVER & Avg. \\
    \midrule
    Standard fine-tuning   & 44.3 & 14.9 & 73.2 & 44.1 & 47.0 & 18.4 & 89.7 & \textbf{51.7} \\
    Top-100 re-ranking     & 44.2 & 14.6 & 75.4 & \textbf{44.7} & \textbf{47.1} & \textbf{18.7} & 88.9 & 51.6 \\
    Query-side fine-tuning & \textbf{45.0} & \textbf{15.0} & \textbf{77.0} & \textbf{45.7} &   44.9 & 17.9 & \textbf{90.2} & 51.0  \\
    Fixed retriever        & 36.8 & 14.5 & 72.0 & 41.1 & 38.0 & 17.7 & 89.3 & 48.3  \\
    \bottomrule
  \end{tabular}
\end{table}

\subsection{Training and evaluating \Atlas}
\label{sec:final}
In this section, we apply the findings from the ablations of the previous sections to train a family of \Atlas~models, ranging from 770M to 11B parameters.
More specifically, we use the Perplexity Distillation objective function, along with the masked language modelling pretext task.
We pre-train these models using a mix of Wikipedia and Common Crawl data, for both the training data and content of the index.
We retrieve 20 documents, and update the index every 2,500 steps and perform re-ranking of the top-100 documents.
We pre-train models for 10,000 iterations using AdamW with a batch size of 128.

\begin{table}[t]
  \centering
  \caption{\textbf{Performance on MMLU as a function of model size.}}
  \label{table:mmlu_model_size}
  \vspace{0.5em}
  \begin{tabular}{l ccc ccc ccc}
    \toprule
    & \multicolumn{3}{c}{5-shot} & \multicolumn{3}{c}{5-shot (multi-task)} & \multicolumn{3}{c}{Full / Transfer} \\
    \cmidrule(lr){2-4} \cmidrule(lr){5-7} \cmidrule(lr){8-10}
                   & 770M &  3B  & 11B  & 770M &  3B  & 11B & 770M &  3B  & 11B \\
    \midrule
    Closed-book T5 & 29.2 & 35.7 & 36.1 &  26.5    &  40.0 & 43.5  & 42.4 & 50.4 & 54.0\\
    \Atlas           & 38.9	&	42.3	&	43.4	&	42.1	&	48.7	&	56.4	&	56.3	&	59.9	&	65.8 \\
\midrule 
 $\Delta$ & +9.8	&	+6.6	&	+7.3	&	+15.6	&	+8.7	&	+12.9	&	+13.9	&	+9.5	&	+11.8\\
    \bottomrule
  \end{tabular}
\end{table}

\subsubsection{MMLU Results} 
As mentioned in section \ref{sec:benchmarks}, we consider four setting for MMLU: 1) a zero-shot setting where we directly apply the pretrained model with no few-shot finetuning  2) a 5-shot setting, where we finetune a model using 5 training examples for each of the 57 domains 3) a 5-shot multitask setting, where, rather than finetuning a model independently for each domain, we train a single  model to perform all tasks and 4) a setting with access to a number of auxiliary datasets, with 95K total training examples.   
We train the models to generate the letter corresponding to the correct answer option (`A', `B', `C' or `D'), and pick the answer with the most likely of the 4 letters at test time. Full technical details can be found in appendix~\ref{app:mmlu}.

\paragraph{Performance vs Parameters.} We start by comparing \Atlas~to closed-book models of different sizes for 5-shot, 5-shot multitask and the full setting, and report results in Table~\ref{table:mmlu_model_size}.
Across these settings, \Atlas~outperforms the closed-book baselines by between 6.6 and 15.6 points, demonstrating consistent utility of retrieval for few-shot language understanding across 57 domains.
The closed-book T5 struggles to perform significantly better than random (25\%) in  few-shot settings with 770M parameters, whereas the equivalent \Atlas{} achieves around 40\%, significantly better than random, despite its small size.
All models improve with more data, but interestingly, the 770M models do not benefit as much from few-shot multitask learning compared to larger models (for closed-book, it actually loses 3 points) suggesting smaller models struggle to grasp the synergies between the tasks in the few-shot setting.
Larger models exploit the multi-task setting well, with \Atlas{} improving more than closed-book. For example, \Atlas{}-11B improves by 13 points  (43.4 $\rightarrow$ 56.4), but equivalent closed-book only improves by 7 (36.1 $\rightarrow$ 43.5).
Finally, on the transfer learning setting, all models improve, but the relative gaps between closed-book at \Atlas{} models remain similar.

\paragraph{De-biasing.} 
When finetuning, we permute which answer option appears with which answer letter to reduce over-fitting and encourage a uniform prior over answer letters. 
However, the model may still exhibit a bias towards some letters, especially in few-shot settings, so we also include a second `de-biased' inference mode in addition the standard inference used above. Here, we run 4 forward passes, one for each cyclic permutation of the answer letter-answer option assignment in the question, e.g. the answer option assigned to letter `A'  becomes `B', what was `B' becomes `C' etc.\footnote{Exploring all answer option permutations would involve 24 forward passes, which improves results by an additional $\sim$1\% over the 4 cyclic permutations, but requires much more compute, so we exclude it here, see Appendix \ref{app:mmlu}}
We then sum the 4 probabilities to obtain the final prediction, which reduces spurious bias towards one of the answer letters (further details in appendix~\ref{app:mmlu}). The results are shown in Table \ref{tab:mmlu_debias_comp}. We find that in zero-shot and 5-shot settings, de-biasing is very effective, improving results by 10.3 and 4.5 points respectively. When more training data is available, the need for de-biasing decreases, leading to only 0.2 point improvement in the multi-task and full data settings.

\paragraph{Comparison to published works} 
Next, we compare our \Atlas{}-11B results with de-biasing  to recently reported results with state-of-the-art large language models such as GPT-3 or Chinchilla, which required significantly more amount of computation to train.
We report results in Table~\ref{table:mmlu_sota}.
We find that \Atlas{} is able to perform significantly better than random in zero-shot, and in conjunction with de-biased inference, achieves zero-shot  scores that exceed 5-shot results reported with GPT3 in the literature (47.1\% vs 43.9\%)~\citep{hendrycks2021mmlu}.
For the 5-shot setting, \Atlas{} outperforms GPT-3 by 4\%, while using 15$\times{}$ less parameters, and 10$\times{}$ less pre-training compute.\footnote{\Atlas{}'s pre-training compute is dominated by the T5 pre-training. The computational requirements for the retrieval-augmented pre-train is orders of magnitude lower}
When multitask-training on the combined 5-shot data, \Atlas{} improves to 56.6\%  close to the 5-shot performance of Gopher (60.0\%).
Finally, on the full data setting, where we train on auxiliary data recommended by the MMLU authors, \Atlas{} reaches an overall accuracy of 65.6\%,  close to the state-of-the-art.
Interestingly, in this setup, \Atlas{} significantly outperforms GPT-3, while on the 5-shot setting, their performance is similar.

\begin{table}[t]
  \centering
  \caption{\textbf{Standard vs de-biased inference for MMLU} These results are reported for \Atlas{}-11B, using cyclic permutations for de-biasing, which increases inference costs by a factor of 4$\times$. }
  \label{tab:mmlu_debias_comp}
  \vspace{0.5em}
  \begin{tabular}{l ccc c}
    \toprule
    &  Zero-shot & 5-shot & 5-shot (multi-task) & Full / Transfer \\
    \midrule
    Standard Inference & 36.8	&	43.4	&	56.4	&	65.8\\
    De-biased Inference & 47.1	&	47.9	&	56.6	&	66.0\\
    \bottomrule
  \end{tabular}
\end{table}

\begin{table}[t]
  \centering
  \caption{\textbf{Comparison to state-of-the-art on MMLU.} $^*$For the 5-shot setting, \Atlas{} uses fine-tuning, while previous works use in-context learning. The \Atlas{} model uses de-biased inference.
    Train FLOPS refers to total the amount of computation necessary to train the model, including pre-training and/or fine-tuning.}
  \label{table:mmlu_sota}
  \vspace{0.5em}
  \small
  \begin{tabular}{ll cc c cccc}
    \toprule
    Setting                 & Model      & Params & Train \textsc{FLOPS} & All  & Hum. & Soc. Sci. & STEM & Other \\
    \midrule
        \multirow{1}{*}{zero-shot}     & \Atlas       & 11B    & 3.5e22 & 47.1 & 43.6&	54.1&	38.0&	54.4  \\
      \midrule
    \multirow{4}{*}{5-shot} & GPT-3      & 175B   & 3.1e23      & 43.9 & 40.8 & 50.4 & 36.7 & 48.8 \\
                            & Gopher     & 280B   & 5.0e23      & 60.0 & 56.2 & 71.9 & 47.4 & 66.1 \\
                            & Chinchilla & 70B    & 5.0e23      & \textbf{67.5} & \textbf{63.6} & \textbf{79.3} & \textbf{55.0} & 73.9 \\
                            & \Atlas$^*$   & 11B    & 3.5e22  & 47.9 & 46.1	&54.6&	38.8&	52.8 \\
      \midrule
     \multirow{1}{*}{5-shot (multi-task)}     & \Atlas       & 11B    &  3.5e22 & 	56.6 & 50.1&	66.4&	46.4&	66.2 \\
    \midrule
    \multirow{3}{*}{Full / Transfer}   & UnifiedQA  & 11B    & 3.3e22 & 48.9 & 45.6 & 56.6 & 40.2 & 54.6 \\
                            & GPT-3      & 175B   & 3.1e23 & 53.9 & 52.5 & 63.9 & 41.4 & 57.9 \\
                            & \Atlas       & 11B    & 3.5e22 & 66.0  & 61.1&	77.2&	53.2&	\textbf{74.4}\\
    \bottomrule
  \end{tabular}
\end{table}

\subsubsection{Open-domain Question Answering Results} 
Next we evaluate \Atlas~on two open-domain question answering benchmarks: NaturalQuestions and TriviaQA.
We compare to prior work, both in a few-shot setting using 64 examples, and using the full training set, and report results in Table~\ref{table:nq_sota}.
On these benchmarks, which require high-degree of memorisation, we clearly see the benefits of retrieval-augmentation.
 \Atlas{}-11B obtains state-of-the-art results on 64-shot question answering, for both NaturalQuestions and TriviaQA.
In particular, it outperforms significantly larger models, such as PaLM, or models that required significantly more training compute such as Chinchilla.
When using the full training set, \Atlas~also obtains state-of-the-art results, for example improving the accuracy on NaturalQuestions from 55.9\% to 60.4\%.
This result is obtained using an index comprised of CCNet and the December 2021 Wikipedia corpora, our default setting for the index.
In section~\ref{sec:temporal} we consider using indexes composed of Wikipedia corpus archived at different dates, and demonstrate an additional +3.6\% on NaturalQuestions when using an index which is temporally matched to NaturalQuestions.
We report performance as a function of model size as well as detailed hyperparameters in Appendix~\ref{app:qa}.

\Atlas{} also compares favorably to recent work exploring retrieval-augmented few-shot question answering with very large models. %
\citet{Lazaridou2022InternetaugmentedLM} explore NaturalQuestions in a 15-shot setup using Gopher, augmenting questions with 50 passages retrieved using Google Search. 
This method consists of generating 4 candidate answers from each retrieved passages, and then re-ranking using either a score inspired by RAG~\citep{lewis2020retrieval} or a more expensive approach.
This method (not shown in our tables) achieves exact match scores of 32.7\% (RAG) and 38.4\% (Ensemble), requiring 50 (RAG) or 450 (Ensemble) forward passes of Gopher-280B per test-time question.
\Atlas{}, using the same 15 training examples and 50 passages achieves 38.7 EM, despite having 25$\times{}$ fewer parameters, and requiring comparatively negligible  compute.

\begin{table}[t]
  \centering
  \caption{\textbf{Comparison to state-of-the-art on question answering.} 
    We report results on NaturalQuestions, and on TriviaQA for both the filtered set, commonly used for open-domain question answering and the unfiltered hidden set for which evaluation is accessible online:~\small\url{https://competitions.codalab.org/competitions/17208}\normalsize. 
    For the 64-shot setting, our model uses fine-tuning, while the other models use prompting.
   }
  \label{table:nq_sota}
  \vspace{0.5em}
  \begin{tabular}{l cc cc cc}
    \toprule
    & \multicolumn{2}{c}{NQ} & \multicolumn{2}{c}{TriviaQA filtered} & \multicolumn{2}{c}{TriviaQA unfiltered}\\
    \cmidrule(lr){2-3} \cmidrule(lr){4-5} \cmidrule(lr){6-7} 
    Model & 64-shot & Full & 64-shot & Full & 64-shot & Full \\
    \midrule
    GPT-3~\citep{brown2020gpt3}      & 29.9 & - &  -   & - & 71.2 & - \\
    Gopher~\citep{rae2021goepher}     & 28.2 & - & 57.2 & - & 61.3 & - \\
    Chinchilla~\citep{hoffmann2022chinchilla} & 35.5 & - & 64.6 & - & 72.3 & - \\
    PaLM~\citep{chowdhery2022palm}       & 39.6 & - &  -   & - & 81.4 & -\\
    RETRO~\citep{borgeaud2021retro}      & - & 45.5 &  -   & - & -    & -\\
    FiD~\citep{izacard2020leveraging}        & - & 51.4 & -    & 67.6 & - & 80.1 \\
    FiD-KD~\citep{izacard2021distilling}     & - & 54.7 &  -   & 73.3 & - & - \\
    R2-D2~\citep{fajcik2021r2d2} & - & 55.9 & -    & 69.9 & - & - \\
    \Atlas       & \textbf{42.4} & \textbf{60.4} & \textbf{74.5} & \textbf{79.8} & \textbf{84.7} & \textbf{89.4} \\
    \bottomrule
  \end{tabular}
\end{table}

\subsubsection{FEVER Results}
We report results on the original 3-class FEVER fact checking test set in Table~\ref{tab:fever}.
We consider a 64-shot setting, with training examples uniformly sampled from the full training set.
Unlike the development and test sets, the train set is imbalanced, with more positive labels than negative, posing a challenge for few-shot learning.
In this setting, we achieve an accuracy of 64.3\%.
We also report a 15-shot setting, with 5 examples uniformly sampled from each class to compare with published results from  Gopher~\citep{rae2021goepher}, where \Atlas{} scores 56.2\%, outperforming Gopher by 5.1 points.
Lastly we fine-tune our model on the full training set, and achieve a score of 78\%, within 1.5\% of the ProoFVer, which uses a specialized architecture, a retriever trained with sentence-level annotations, and is supplied with the Wikipedia corpus released with FEVER, whereas \Atlas{} retrieves from CCNet and the December 2021 Wikipedia dump. 
If we give \Atlas{} an index comprised of the FEVER Wikipedia corpus, we set a new state-of-the-art of 80.1\%

\begin{table}[t]
  \centering
  \caption{\textbf{Comparison to state-of-the-art on FEVER.} We report accuracy on FEVER test set, for which evaluation is available here:~\small\url{https://competitions.codalab.org/competitions/18814}\normalsize.
    For the few-shot settings, our model uses fine-tuning while other models use prompting. $^\dagger$uses an index composed of the FEVER Wikipedia corpus.
   }
  \label{table:fever}
  \vspace{0.5em}
  \begin{tabular}{lccc}
    \toprule
    & 15-shot & 65-shot & Full dataset\\
    \midrule
    Gopher\small~\citep{rae2021goepher}\normalsize & 51.1 &- &- \\
    ProoFVer\small~\citep{krishna2021proofver}\normalsize& - &- & 79.5 \\
    \Atlas{} & \textbf{56.2} & \textbf{64.3} & 78.0 / \textbf{80.1}$^{\dagger}$ \\
    \bottomrule
  \end{tabular}
\label{tab:fever}
\end{table}

\subsubsection{KILT Results}
Finally we evaluate \Atlas{} on KILT, a benchmark composed of several different knowledge intensive tasks, which was described in section \ref{sec:benchmarks}.
We report results on test sets in Table~\ref{tab:kilt_sota} for which evaluation is available online\footnote{\url{https://eval.ai/web/challenges/challenge-page/689}}.
The KILT versions of datasets are filtered, and thus results for datasets we have evaluated elsewhere are not directly comparable on KILT (i.e. FEVER, NQ and TQA).  
We consider both a 64-shot setting and a full fine-tuning setting, in both cases we train \Atlas{} individually on each dataset.
More details on the hyperparameters and development set results are reported in Appendix~\ref{app:kilt}.
For 64-shot, we greatly exceed random performance, and are even competitive with some fully-finetuned models on the leaderboard, such as for FEVER, where our 64-shot \Atlas{} is only 2-2.5 points behind Sphere, SEAL and Re2G, and outperforms Sphere and SEAL on zero-shot RE.
In the full dataset setting, \Atlas{} is within 3\% to the state-of-the-art for 3 datasets, and sets the state-of-the-art in the remaining five datasets.

\begin{table*}[t]
    \caption{\textbf{Downstream results on the KILT hidden test sets} Downstream metrics are accuracy (AIDA CoNLL-YAGO, FEVER, T-REx, zero-shot RE), exact match (Natural Questions, HotpotQA, TriviaQA), or F1 (Wizard of Wikipedia).}
    \label{tab:kilt_sota}
    \vspace{0.5em}
    \centering
    \scalebox{0.98}{
        \begin{tabular}{lcccccccccc}
    \toprule
    \multirow{2}{*}{Model}         & {AIDA} & {FEV} & {T-REx} & {zsRE} & {NQ} & {HoPo} & {TQA}  & {WoW}  \\
    & {\textsc{acc}} & {\textsc{acc}} & {\textsc{acc}} & {\textsc{acc}} & {\textsc{em}}   & {\textsc{em}}   & {\textsc{em}} & {\textsc{f1}}  \\ \midrule
    GENRE~\citep{cao2021autoregressive} & 89.9 & {-} & {-} & {-} & {-} & {-} & {-} & {-} \\
    Sphere~\citep{piktus2021sphere} & {-} & 89.0 & 81.7  & 74.2 & 51.6 & 38.3 & 72.7 &  15.5 \\
    SEAL~\citep{bevilacqua2022seal} & {-} & 89.5 & 83.6  & 74.6 & 53.7 & 40.5 & 70.9 & 18.3 \\
    Re2G~\citep{glass2022re2g} & {-} & 89.6 & \textbf{87.7} & {-} & 51.7 & {-} & 76.3 & 18.9 \\
    FID with RS \citep{Hofsttter2022MultiTaskRT} &  {-}  & {92.2}&   85.2 &  \textbf{83.7}&  {61.2}&  {39.1}&  \textbf{84.6}&   {20.6}\\
    \midrule
    \Atlas{}, 64-shot & 66.5 & 87.1 & 58.9 & 74.9 & 43.6 & 34.7 & 76.4 & 15.5 \\
    \Atlas{}, full train set & \textbf{90.6} & \textbf{93.5} & 85.1 & 80.8 & \textbf{61.3} & \textbf{50.6} & 84.0 & \textbf{21.6} \\
    \bottomrule
    \end{tabular}
    
    }
\end{table*}

\section{Analysis}

\subsection{Interpretability and Leakage}
\begin{figure}[t]
\centering
\includegraphics[width=\textwidth]{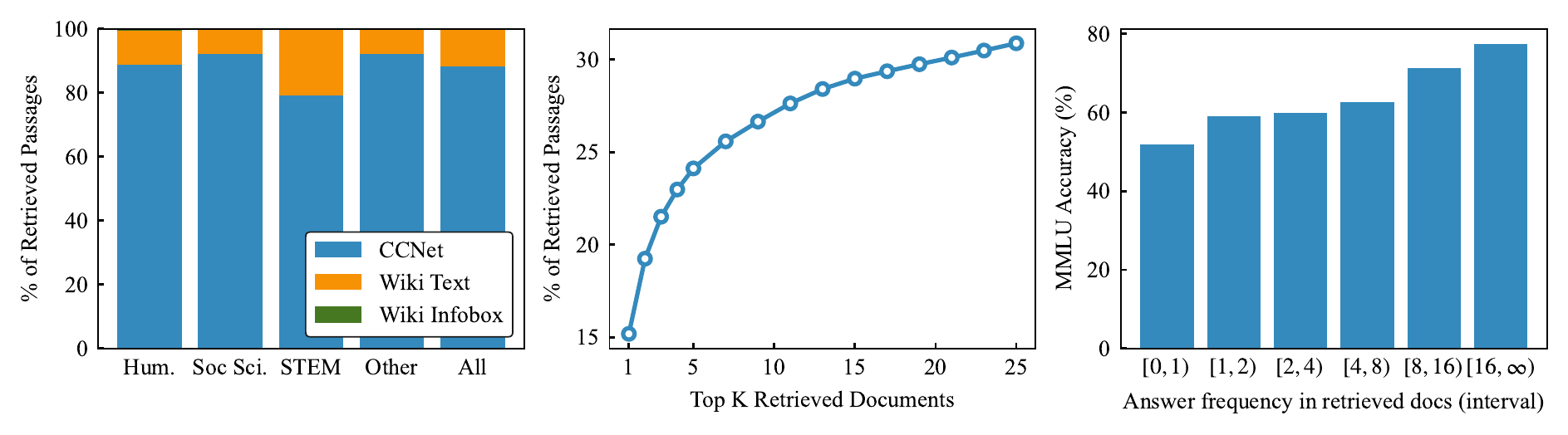}
\caption{\textbf{MMLU Retrieval Analysis. Left}: Fraction of sources of top 30 retrieved passages for MMLU  from CCNet, Wikipedia passages and info boxes for the 5-shot multitask \Atlas{}. \textbf{Center}: How often the text of the correct MMLU answer option appears in retrieved passages, as a function of the number of retrieved passages. \textbf{Right}: MMLU accuracy as a function of answer occurrence frequency in retrieved passages set}
\label{fig:mmlu_retrieval_anslysis}
\end{figure}

An advantage of semi-parametric models like \Atlas{} is the ability to inspect retrieved items to aid interpretability.
To better understand how well \Atlas{} retrieves, and how it uses retrieved passages, 
we examine the retrieved passages for multi-task few-shot MMLU.
As shown in the left panel of Figure \ref{fig:mmlu_retrieval_anslysis}, the model retrieves the majority of its passages from CCNet (85\% on average).
Wikipedia makes up about 15\% of retrieved passages, which is higher than we would expect under a uniform prior, given Wikipedia only makes up about 10\% of the index.
The fraction of Wikipedia retrieval varies between MMLU domains, with the model using Wikipedia to a greater extent for STEM domains, and least for social sciences.
The domain making the greatest use of Wikipedia is ``abstract algebra'' (73\%), and the least is ``moral scenarios'' (3\%). 
We also note that the MMLU-finetuned \Atlas{} does not make significant use of Wikipedia infobox passages.

We can also analyse the content of passages to assess how they may useful for accomplishing the downstream task. 
The middle panel of Figure \ref{fig:mmlu_retrieval_anslysis} shows how often retrieved documents contain the text of the correct answer option. There being at least one mention of the correct answer choice in 30\% of test questions in the top 25 passages.
\footnote{Note: Depending on the question, it may not be important or useful to retrieve the exact text of the answer in MMLU, and as such, a hits@k value of 30\% does not imply that retrieval fails to surface useful information in 70\% of cases}
The right panel shows that the accuracy on MMLU increases when the correct answer option text occurs more frequently in retrieved passages, rising from 55\% for questions when the answer option does not appear, to 77\% for questions mentioned more than 15 times.

A human analysis of retrieved documents revealed that documents are helpful for answering questions in a number of different ways.
Manual inspection of a sample of 50 correctly-answered questions revealed that 44\% contained at least partially useful background information.
These are documents that would improve the likelihood of a non-expert human answering correctly, such as contextual clues surrounding a quotation from a question, or helpful numerical figures for quantity-based questions, which help to narrow down the answer options to a smaller range.
In a further 26\%  of cases, a passage contained all the necessary information to answer the question, stated in a straightforward way.
If read competently, such passages make the question simple to answer, and often include information such as canonical definitions, or the exact numerical answer requested in the question.
28\% of retrieval sets did not contain obvious information which would make the question easier.
Finally, 2\% contained the verbatim question in a passage, together with its answer.

Given that MMLU has been created from pre-existing exams, it is possible that these questions appear on the open web.
Models trained on web data (or, in our case, retrieving from it) run the risk of answering correctly not through generalisation, but by verbatim memorisation, which could lead to misleadingly high scores.
In some very large language models, which can verbatim memorize and recall large parts of their pre-training data~\citep{DBLP:conf/uss/CarliniTWJHLRBS21},  efforts have sometimes been made to filter occurrences of downstream instances from pre-training data, but this has not been performed for MMLU in the literature.
In order to assess the prevalence of MMLU leakage in our index, we manually checked retrieval results for questions where the longest n-gram overlap between the question (without answer options) and a passage was at least 75\% the length of the question.
This resulted in an estimate of leakage of 2.8\% of questions from our CC-Net corpus.

A benefit of retrieval-augmented models such as \Atlas{} is the editability of its knowledge (see section \ref{sec:temporal} for additional analysis). 
To estimate pure, non-leaked performance, we can filter out any potentially-leaked passages from retrieved results and rerun the language model.
The MMLU score drops slightly when controlling for this leakage from 56.4 to 55.8\% (-.5\%).
We note that our CC-net corpus is relatively small compared to the pre-trained corpora of recent very large models, which are trained on up to 1.4 trillion tokens~\citep{hoffmann2022chinchilla}, 35x the size of our index, making it likely that models trained on corpora of that size would observe more MMLU leaked examples, but detecting such leakage is challenging in non-retrieval augmented models.

\subsection{Temporal Sensitivity and Updateability}
\label{sec:temporal}
A benefit of retrieval-augmented models is that they can be kept up-to-date without  retraining, by updating or swapping their index at test time.
To assess the effectiveness of this mechanism in \Atlas{},  we first construct a dataset of time-sensitive questions derived  from TempLAMA~\citep{dhingra-etal-2022-time}.
TempLAMA is a collection of templated cloze questions derived from Wikidata and Wikidata where the correct answer changes over time.
We select a subset of questions from this dataset which have a different answer in 2017 and 2020,  for example, \texttt{Question:} \textit{Theo Walcott plays for \_\_\_} \texttt{Answer:} \textit{Arsenal F.C. (2017), Everton F.C. (2020)}, and form a small training set of 248 training, 112 development and 806 test questions.

Using this dataset, we finetune closed-book T5-XXL and \Atlas{} using the questions and the 2017 answers, supplying \Atlas{} with a 2017 Wikipedia index, and then measure exact match accuracy on the 2017 test set.
The results can be found in the first row and first two columns of Table~\ref{table:templama}.
We first observe that, as expected, \Atlas{} greatly outperforms T5 (57.7\% c.f. 12.1\%).
We also note that, as desired, both T5 and \Atlas{} almost never generate an answer from 2020 when trained with the 2017 answers, scoring 2.8\% and 1.5\% respectively (first row, second two columns of Table~\ref{table:templama}).
However, as shown in row 2, we can swap the \Atlas{} index to a 2020 Wikipedia index,  \emph{without retraining}, and find that \Atlas{} updates its predictions accordingly, with 2020 accuracy rising to a similar level to its 2017 performance (53.1\%), whereas the purely parametric T5 has no such updateability mechanism.

This demonstrates that \Atlas{} can be faithful and condition strongly on its supplied index.
Furthermore, this zero-shot updateability mechanism has the useful property of staying up-to-date without requiring up-to-date annotated  data, or continuous, lifelong pre-training, as would be may required for a large parametric-only model. 
Rows 3 and 4 of Table \ref{table:templama} complete the picture, where this time we train with 2020 answers, and demonstrate \Atlas{} can zero-shot transfer backwards in time to 2017 effectively too (50.1\%). 
Interestingly, T5 is unable answer questions from 2020 well, even when trained with 2020 answers (3.6\%), likely because it was pre-trained on data pre-dating 2020~\citep{dodge-etal-2021-documenting}.

\begin{table}[t]
  \begin{center}
   \caption{\textbf{Results on our TempLAMA-derived dataset.} We report performance for a static, closed-book T5-11B, as well as \Atlas{}-11B supplied with a test-time Wikipedia index from 2017 or 2020. We evaluate models finetuned on a small training set of 248 time-sensitive cloze-question-answer pairs, using answers either from 2017 or 2020.
  Good models should score highly when the test set year matches the year of the test-time index, and score low otherwise.}
      \vspace{0.5em}

  \begin{tabular}{l c cc cc}
    \toprule
     && \multicolumn{2}{c}{2017 Test Set Acc.} & \multicolumn{2}{c}{2020 Test Set Acc.} \\
    \cmidrule(lr){3-4} \cmidrule(lr){5-6} 
    Train Set & Test-time Index & Closed-book & \Atlas &Closed-book & \Atlas  \\
    \midrule
\multirow{2}{*}{2017 answers} & 2017 & 12.1 & 57.7 & 2.9 & 1.5\\
& 2020 & 12.1 & 10.2 & 2.9 & 53.1 \\
    \midrule
\multirow{2}{*}{2020 answers} & 2017 & 4.8

& 50.1 & 3.6& 4.2\\
& 2020 & 4.8
& 3.5 & 3.6 & 60.5\\
    \bottomrule
  \end{tabular}
   
  \label{table:templama}
  \end{center}
\end{table}

We also examine temporal effects for NaturalQuestions. 
NaturalQuestions is a dataset composed of search queries collected via the Google search engine in a short period of time.
Thus data have a strong temporal bias, with a lot of questions about the 2018 World Cup for example.
Moreover some questions are ambiguous without specification of the temporal context.
For instance, for the question \textit{``when did ireland last beat england at twickenham''}, the expected answer is 2018 in NaturalQuestions, while Ireland also beat England at Twickenham in 2022 as well as many other times before.
In Table~\ref{table:nq_index}, we report results obtained by finetuning \Atlas{} using different Wikipedia dumps for the index.
We observe that the 2018 December Wikipedia dump, which is close to the date of data collection, leads to the best results for both few-shot and full fine-tuning.
In particular, it leads to a new state-of-the-art of 64~EM on NaturalQuestions.

\begin{table}[t]
  \centering
      \caption{\textbf{Impact of index data temporality on NaturalQuestions.}
    We report exact match performance on NaturalQuestions using different Wikipedia dumps in the index.
    We observe that the dump from December 2018, commonly used for NaturalQuestions, leads to the best result.}
        \vspace{0.5em}

  \begin{tabular}{l ccccc}
    \toprule
    & Dec. 2017 & Dec. 2018 & Aug. 2019 & Dec. 2020 & Dec. 2021 \\
    \midrule
    64-shot & 44.7 & \textbf{45.1} & 44.1 & 44.0 & 41.3 \\
    Full    & 63.2 & \textbf{64.0} & 62.4 & 61.1 & 59.6 \\
    \bottomrule
  \end{tabular}

  \label{table:nq_index}
\end{table}

\subsubsection{Index Compression}
\begin{figure}[t]
    \centering
    \includegraphics[width=0.85\textwidth]{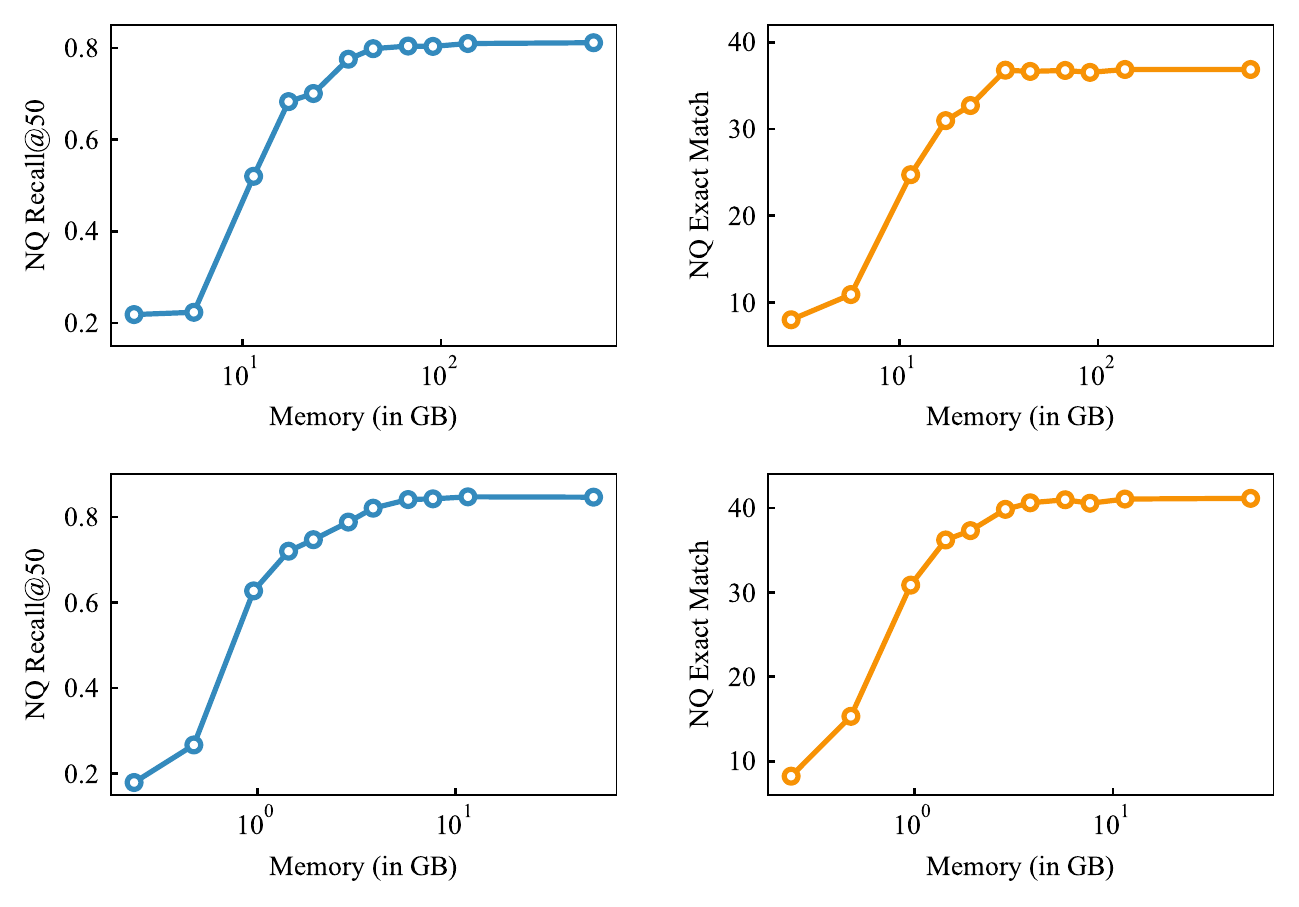}
\caption{\textbf{Index Compression:} \Atlas{}-3B 64-shot NQ performance (left column: Retrieval Recall@50, right column: QA Exact Match score), as a function of index size, for different levels of quantisation.
    The right-most point in each plot represents the uncompressed index.
  \textbf{Top Row:} Wikipedia + CC Index. \textbf{Bottom Row:} Wikipedia Index.
}
    \label{fig:compressed_indexes}
\end{figure}

Maintaining dense retrieval indices can be memory-intensive, especially as the number of indexed items is scaled.
In this section, we briefly analyse the memory requirements of \Atlas{}'s index in the case of a) a Wikipedia index and b) the combined  CC and Wikipedia index used in most of the experiments above.

There are two sources of memory pressure for \Atlas{}'s retrieval component -- the passages themselves, and the document embedding index.
The tokenized passages, once binarized, require 11GB and 130GB of storage for the Wikipedia and combined indices respectively. 
These passages do not need to be stored in expensive GPU RAM, and could even be memory-mapped to disk, sharded across nodes or compressed if required, and thus do not represent a limiting hardware challenge in this context.
The embedding index itself, however, must be stored in GPU RAM for fast search, and thus its size is more sensitive.
In the above experiments, we perform exact search over our index, which is achieved by sharding the index over all the the available GPUs, and computing the search in parallel.
The index is stored at fp16 precision, resulting in a total GPU memory requirement of 49 GB and 587 GB for the Wikipedia and combined indices, respectively.

This large GPU memory requirement for the index limits accessibility and ease of deployment.
However, many index compression techniques are available for nearest neighbour search, which can often dramatically reduce memory requirements at the cost of some retrieval accuracy.
Following \citet{izacard2020memory}, we explore the effect of Product Quantization \cite[PQ,][]{jegou_quantization}, a popular lossy compression technique on \Atlas{}-3B's accuracy for the 64-shot NQ task at different compression levels.

The results are shown in Figure \ref{fig:compressed_indexes}.
We find that substantial compression is possible before the onset of significant performance degradation. 
Namely, the Wikipedia index can be compressed from 49GB to 4GB with negligible drop in retrieval precision and exact match.
Likewise, the combined index can be compressed from 587GB to 50GB without serious degradation, indicating that the combined index could be loaded onto a single 80GB GPU.

\section{Discussion}
In this paper, we introduce \Atlas, a large retrieval-augmented language model.
By jointly pre-training the retriever module and the language model, we show that \Atlas{} has strong few-shot learning capabilities on a wide range of knowledge intensive tasks, including NaturalQuestions, TriviaQA, FEVER, 8 KILT tasks and 57 MMLU tasks.
For example, \Atlas{}-11B  reaches more than 42\% accuracy on NaturalQuestions and 84.7\% on TriviaQA when training on 64 examples, which is an improvement of almost 3 points compared to PaLM, a 540B parameters model, which required 50x more pre-training compute.
We also provided detailed ablations and analyses for what factors are important when training such retrieval-augmented models, and demonstrated \Atlas{}'s updateability, interpretability and controlability capabilities.
Lastly, we demonstrated that \Atlas{} is also powerful in full-dataset settings obtaining a new state-of-the-art results on NaturalQuestions, TriviaQA, FEVER, and 5 KILT tasks.




\bibliography{references}
\bibliographystyle{tmlr}

\clearpage
\appendix
\section{Training details and additional results}

\subsection{MMLU}
\label{app:mmlu}
\subsubsection{Training Details}

\paragraph{Featurization} MMLU consists of multiple choice questions with four possible lexicalized answer options.
We represent the input using the following template: 

\texttt{question: \{question text\} \\
options: (A) \{answer 1 text\} (B) \{answer 2 text\} (C) \{answer 3 text\} (D) \{answer 4 text\}\\ answer: [MASK\_0]
}

and train the model to generate the mask token followed by the letter of the correct answer:

\texttt{[MASK\_0] \{correct answer option letter\}}

This format closely matches the format of MLM pre-training objective, aiding few-shot learning. When training, we permute the order of the answer options, i.e. shuffling which answer option appears as letter A etc.
This helps reduce overfitting, and encourages a uniform prior on the letters.

\paragraph{Standard inference} Once trained we obtain predictions from the model by selecting the pre-softmax logits for the tokens A, B, C and D, and performing a softmax over them to obtain a distribution over the 4 answer options.
For standard inference, we then simply return the answer corresponding to the argmax of this distribution.

\paragraph{De-biased Inference} As mentioned in the main text, even though our model is finetuned with data that encourages a uniform prior over answer letters (by permuting which answer option letter is used with which lexical answer option text in training data), this may not be enough to ensure the model has no residual bias towards specific letters.
Consider answers $a$, questions $q$ and a nuisance variable $z \in \mathcal{Z}$, which represents the ordering of the answer options or, equivalently, which answer letter gets assigned to which answer option text.
There are 4 answer options in MMLU, and thus $|\mathcal{Z}| = 24$ unique ways they can be ordered, or assigned to given letters.
Running our model with our standard inference for a question $\textsf{q}$, corresponds to calculating $p(a|q=\textsf{q},z=\textsf{z})$ for the answer ordering $\textsf{z}$ that happens to appear in the dataset.
We can control for $z$ by running the model with all possible answer orderings in the input, and marginalizing: $p(a|q=\textsf{q}) = \sum_{z' \in \mathcal{Z}} p(a|q=\textsf{q},z=z') p(z=z'|q=\textsf{q})$, and assuming  $p(z=z'|q=\textsf{q})$ is uniform (no answer ordering is more likely than another), this reduces to simply $p(a|q=\textsf{q}) \propto \sum_{z' \in \mathcal{Z}} p(a|q=\textsf{q},z=z')$. This procedure requires 24 forward passes, one for each answer ordering, so is 24$\times$  slower than standard inference. 
Table \ref{tab:mmlu_debiasing} shows the result of applying the full permutation de-biasing, which leads to an 12\% improvement zero-shot and 6\% in 5-shot performance overall.   
Empirically, using only the cyclic permutations of the answer order provided in the original dataset (of which there are 4) works nearly as well, which is what we report in the main paper, and only increases inference compute by a factor of 4, rather than 24. Cyclic permutation de-biasing improves over standard inference by 10\% in zero-shot and 5\% in 5-shot.
Empirically, de-biased inference is largely unnecessary when training in the 5-shot multitask or full dataset setting, as there is enough data for the model to learn a more uniform prior over the letters.

\begin{table}[t]
  \centering
  \caption{MMLU scores with de-biasing:}
  \label{tab:mmlu_debiasing}
  \vspace{0.5em}
  \small
  \begin{tabular}{ll c cccc}
    \toprule
    Setting& Model  & All  & Hum. & Soc. Sci. & STEM & Other \\
    \midrule
\multirow{3}{*}{zero-shot}
& Standard &36.8	&37.5&	39.0&	30.2&	39.7\\
&All permutations& 48.5&	45.7&	55.2&	39.4&	54.4\\
& Cyclic Permutations& 47.1&	43.6&	54.1&	38.0&	54.9\\
    \midrule
    \multirow{3}{*}{5-shot}
& Standard & 43.4 &	41.8&	49.3&	33.9&	48.8    \\
& All permutations & 49.0&	46.0&	56.1&	40.5&	54.6    \\
& Cyclic Permutations & 47.9&	46.1&	54.6&	38.8&	52.8    \\

    \bottomrule
  \end{tabular}
\end{table}

\paragraph{Evaluation} We evaluate by following the method of \citet{hendrycks2021mmlu}, namely, micro-averaging across all 57 domains to obtain overall accuracy. 
We quote the results of GPT3~\citep{brown2020gpt3} and UnifiedQA~\citep{khashabi-etal-2020-unifiedqa} from the MMLU leaderboard at \url{https://github.com/hendrycks/test}. For Chinchilla and Gopher, we calculate the scores on the categories using the full MMLU results from \citet{hoffmann2022chinchilla}.

\paragraph{Index} The index used for MMLU for all MMLU experiments in the main paper comprised of concatenation of the Wikipedia passages, Wikipedia info boxes and Common Crawl indices, for a total of 387M passages. We can assess the importance of the index by running a model without the common crawl data, leading to a 5-shot multitask result of 52.8\%, compared to 56.4\% for the full model, a drop of 3.6\%.
This indicates that whilst the Wikipedia data is sufficient do well on the task, the addition of the CC data improves results further.

\paragraph{Hyperparameters and development data} Selecting hyperparameters is challenging in few-shot settings.
We do not assume access to an in-domain development set for the 5-shot task. Instead, we determine a set of hyperparameters for the 5-shot task using data from RACE, one of the auxiliary datasets provided by MMLU. Here, we sample 5 sets of 5-shot training data, and for each model size, we explore batch size $\{32, 64\}$,  learning rates for the language model and retriever \{(5e-5, 1e-5), (4e-5, 4e-5)\}, retriever temperature $\{0.1, 0.01\}$ and a fixed number of training steps $\{16, 32, 64, 128\}$, picking the setting that achieves strongest RACE validation scores.
Having determined these hyperparameters, we apply them directly to the 5-shot MMLU task.
For the 5-shot multi-task and full/transfer settings, we use the same batch size, temperatures and learning rates as the 5-shot task, but use a set of 285 MMLU validation examples (5 per domain) in order to determine the total number of training steps and for early stopping.
The hyperparameters selected in the MMLU experiments can be found in table \ref{tab:mmlu_hps}.
We use query-side finetuning for the 5-shot and 5-shot multitask settings, and top-128 reranking for the full setting. For all MMLU runs we retrieve 30 documents

\begin{table}[t]
  \centering
  \caption{Hyperparameters for MMLU}
  \label{tab:mmlu_hps}
  \vspace{0.5em}
  \begin{tabular}{l ccc}
    \toprule
& 770M &  3B  & 11B  \\
\midrule
Batch size & 64 & 64 & 64\\
Learning rate & (5e-5, 1e-5)&(5e-5, 1e-5)&(5e-5, 1e-5) \\
Retriever Temperature & 0.1 & 0.1 & 0.1  \\
5-shot train steps & 64 & 32 & 16\\
5-shot (multitask) max train steps & 2000 & 500 & 250 \\
Full / transfer max train steps & 5000 & 2000 & 2000\\
    \bottomrule
  \end{tabular}
\end{table}

\paragraph{Inter-run Variance} few-shot learning is well-known to suffer from high variance.
In the main paper, we quote the result obtained with our first run.
In order to  assess the effect of noise and variance, we ran the 5-shot experiment with \Atlas{} 5 times.\footnote{This experiment was performed with a slightly different index to the main experiments, which achieves a stronger result}
We observe high variance for individual domains, sometimes as high as 20\%, however, once aggregated across all 57 domains, the inter-run variance is low.
The overall scores for these different runs, when using the same hyperparameters are shown in table \ref{tab:mmlu_var}. Due the effects of averaging over the many domains that comprise MMLU, the inter-run variance is quite modest on the aggregated metrics, with a std deviation of 0.5 in this experiment.

\begin{table*}
    \caption{
    Interrun Variance for 5-shot MMLU using \Atlas{}-11B}
    \label{tab:mmlu_var}
    \vspace{0.5em}
    \centering
 \begin{tabular}{c ccccc}
    \toprule
Run \#  & All  & Hum. & Soc. Sci. & STEM & Other \\
    \midrule
1 & 45.2&	40.6&	54.1&	37.1&	51.1\\
2 & 45.1&	39.8&	54.4&	37.1&	52.0\\
3 & 45.0&	40.0&	54.1&	37.7&	51.1\\
4 & 45.6&	41.3&	54.7&	37.0&	51.6\\
5 & 44.3&	40.6&	50.7&	38.1&	49.8\\
\midrule
Ave: & $45.0\pm0.5$ & $40.5\pm0.6$& $53.6\pm1.6$&	$37.4\pm0.5$ &	$51.1\pm0.8$\\
    \bottomrule
  \end{tabular}
\end{table*}

\paragraph{Closed-Book Baselines} The closed book baselines we compare \Atlas{} to in table \ref{table:mmlu_model_size} are initialized from the same T5 model as their respective \Atlas{}, and then pre-trained with MLM for the same number of steps (10K) using the same pre-training data as \Atlas{}, for fairer comparison. The same procedure as for \Atlas{} was used to determine hyperparameters for MMLU for the closed-book model.s

\subsubsection{Full results}

Tables \ref{tab:full_mmlu} and \ref{tab:full_mmlu_cb} shows the full MMLU scores for each domain for \Atlas{} and  the closed book T5 respectively. The full results for the cyclic-permutation-de-biased \Atlas{}-XXL can be found in Table \ref{tab:full_mmlu_debias}.

\begin{table*}
    \centering
  \vspace{0.5em}
        \scalebox{0.79}{
    \begin{tabular}{r ccc ccc ccc}
    \toprule
    & \multicolumn{3}{c}{5-shot} & \multicolumn{3}{c}{5-shot (multi-task)} & \multicolumn{3}{c}{Full / Transfer} \\
    \cmidrule(lr){2-4} \cmidrule(lr){5-7} \cmidrule(lr){8-10}
& 770M &  3B  & 11B  & 770M &  3B  & 11B & 770M &  3B  & 11B \\
All	&	38.9	&	42.3	&	43.4	&	42.1	&	48.7	&	56.4	&	56.3	&	59.9	&	65.8	\\
\midrule
Humanities	&	37.3	&	40.0	&	41.9	&	37.7	&	46.4	&	50.0	&	50.9	&	53.0	&	60.3	\\
Social Sciences	&	41.7	&	46.8	&	49.3	&	47.5	&	53.7	&	65.6	&	66.0	&	70.8	&	77.2	\\
STEM	&	32.3	&	35.0	&	33.9	&	34.4	&	39.4	&	46.2	&	44.8	&	50.7	&	53.4	\\
Other	&	44.9	&	48.1	&	48.8	&	50.4	&	55.9	&	66.6	&	65.5	&	68.1	&	74.4	\\
\midrule
abstract algebra	&	30.0	&	27.0	&	28.0	&	27.0	&	31.0	&	30.0	&	22.0	&	27.0	&	33.0	\\
anatomy	&	28.9	&	50.4	&	45.2	&	44.4	&	57.8	&	64.4	&	57.8	&	68.9	&	69.6	\\
astronomy	&	55.3	&	59.9	&	59.2	&	52.6	&	66.4	&	67.8	&	69.1	&	78.3	&	79.6	\\
business ethics	&	49.0	&	51.0	&	48.0	&	50.0	&	62.0	&	60.0	&	51.0	&	70.0	&	68.0	\\
clinical knowledge	&	41.9	&	44.9	&	40.0	&	46.8	&	54.3	&	64.9	&	64.2	&	72.5	&	74.0	\\
college biology	&	38.2	&	45.8	&	50.0	&	36.8	&	52.1	&	63.2	&	63.2	&	72.2	&	78.5	\\
college chemistry	&	32.0	&	29.0	&	29.0	&	31.0	&	33.0	&	38.0	&	45.0	&	39.0	&	45.0	\\
college computer science	&	33.0	&	35.0	&	30.0	&	23.0	&	29.0	&	30.0	&	43.0	&	48.0	&	47.0	\\
college mathematics	&	31.0	&	31.0	&	28.0	&	29.0	&	27.0	&	34.0	&	32.0	&	29.0	&	36.0	\\
college medicine	&	31.2	&	35.8	&	38.2	&	50.3	&	40.5	&	52.0	&	60.1	&	59.5	&	63.6	\\
college physics	&	20.6	&	26.5	&	31.4	&	21.6	&	28.4	&	39.2	&	27.5	&	44.1	&	42.2	\\
computer security	&	53.0	&	50.0	&	55.0	&	49.0	&	61.0	&	64.0	&	69.0	&	71.0	&	76.0	\\
conceptual physics	&	34.9	&	41.7	&	37.4	&	40.9	&	43.4	&	57.0	&	53.2	&	58.3	&	59.6	\\
econometrics	&	28.9	&	21.1	&	27.2	&	26.3	&	25.4	&	34.2	&	28.9	&	37.7	&	36.8	\\
electrical engineering	&	26.9	&	31.7	&	31.7	&	38.6	&	44.1	&	51.7	&	61.4	&	60.7	&	67.6	\\
elementary mathematics	&	25.9	&	28.8	&	29.4	&	29.6	&	30.2	&	32.8	&	29.6	&	35.5	&	33.9	\\
formal logic	&	34.9	&	33.3	&	33.3	&	23.0	&	30.2	&	29.4	&	34.1	&	38.9	&	34.1	\\
global facts	&	28.0	&	34.0	&	34.0	&	36.0	&	40.0	&	49.0	&	50.0	&	49.0	&	52.0	\\
high school biology	&	24.8	&	37.7	&	27.7	&	48.7	&	57.1	&	66.5	&	66.5	&	76.8	&	81.9	\\
high school chemistry	&	34.5	&	31.0	&	31.0	&	31.5	&	36.5	&	48.3	&	44.8	&	52.2	&	52.2	\\
high school computer science	&	31.0	&	39.0	&	28.0	&	37.0	&	42.0	&	42.0	&	50.0	&	59.0	&	57.0	\\
high school european history	&	42.4	&	49.7	&	53.3	&	50.9	&	58.2	&	69.7	&	70.9	&	73.9	&	80.0	\\
high school geography	&	38.9	&	42.4	&	50.0	&	46.5	&	56.6	&	69.2	&	74.2	&	80.8	&	82.8	\\
high school gov. and pol.	&	57.5	&	60.6	&	60.1	&	52.9	&	64.8	&	76.7	&	80.8	&	85.5	&	91.7	\\
high school macroeconomics	&	32.8	&	39.7	&	44.9	&	39.0	&	45.6	&	57.2	&	55.1	&	63.1	&	66.7	\\
high school mathematics	&	30.7	&	33.0	&	35.6	&	28.1	&	27.8	&	37.0	&	30.7	&	34.8	&	37.0	\\
high school microeconomics	&	34.5	&	42.9	&	45.4	&	44.1	&	51.7	&	68.9	&	63.4	&	70.2	&	81.1	\\
high school physics	&	18.5	&	24.5	&	22.5	&	25.8	&	25.8	&	33.1	&	27.2	&	30.5	&	39.7	\\
high school psychology	&	52.8	&	61.1	&	59.8	&	56.7	&	67.2	&	79.4	&	76.3	&	84.0	&	87.0	\\
high school statistics	&	39.8	&	29.6	&	34.7	&	27.3	&	34.7	&	38.0	&	37.0	&	43.1	&	45.8	\\
high school us history	&	43.6	&	49.0	&	55.9	&	46.1	&	57.8	&	59.8	&	62.7	&	72.5	&	76.5	\\
high school world history	&	48.1	&	52.7	&	59.9	&	48.1	&	66.2	&	65.4	&	70.0	&	78.5	&	79.7	\\
human aging	&	46.2	&	44.8	&	39.5	&	48.0	&	55.2	&	60.1	&	56.1	&	68.2	&	73.1	\\
human sexuality	&	41.2	&	43.5	&	27.5	&	46.6	&	51.1	&	59.5	&	77.1	&	72.5	&	81.7	\\
international law	&	54.5	&	57.9	&	60.3	&	55.4	&	72.7	&	73.6	&	81.8	&	82.6	&	85.1	\\
jurisprudence	&	38.9	&	55.6	&	32.4	&	53.7	&	60.2	&	73.1	&	76.9	&	73.1	&	81.5	\\
logical fallacies	&	43.6	&	54.0	&	57.1	&	44.2	&	58.3	&	70.6	&	64.4	&	73.0	&	76.7	\\
machine learning	&	36.6	&	34.8	&	28.6	&	31.3	&	37.5	&	46.4	&	36.6	&	47.3	&	50.9	\\
management	&	45.6	&	51.5	&	52.4	&	48.5	&	52.4	&	81.6	&	78.6	&	75.7	&	87.4	\\
marketing	&	59.4	&	67.1	&	70.5	&	66.7	&	74.4	&	83.8	&	83.8	&	83.3	&	91.9	\\
medical genetics	&	50.0	&	53.0	&	58.0	&	56.0	&	61.0	&	75.0	&	68.0	&	78.0	&	81.0	\\
miscellaneous	&	63.0	&	64.2	&	68.8	&	64.0	&	72.4	&	84.3	&	85.4	&	83.9	&	90.9	\\
moral disputes	&	37.0	&	41.3	&	41.3	&	40.8	&	50.3	&	60.1	&	61.9	&	66.2	&	73.7	\\
moral scenarios	&	24.7	&	24.7	&	26.5	&	21.9	&	26.9	&	26.6	&	23.8	&	23.8	&	35.8	\\
nutrition	&	40.9	&	45.1	&	45.1	&	49.0	&	52.3	&	67.0	&	64.7	&	68.6	&	76.8	\\
philosophy	&	48.6	&	50.5	&	56.3	&	49.8	&	59.2	&	69.5	&	70.4	&	73.0	&	77.8	\\
prehistory	&	45.7	&	50.0	&	52.8	&	54.9	&	64.8	&	74.4	&	69.8	&	75.0	&	80.6	\\
professional accounting	&	28.4	&	33.0	&	34.0	&	35.1	&	34.0	&	45.7	&	43.6	&	46.1	&	51.8	\\
professional law	&	32.4	&	33.5	&	34.8	&	30.4	&	37.6	&	39.1	&	41.5	&	41.5	&	50.5	\\
professional medicine	&	29.4	&	26.1	&	27.6	&	34.6	&	40.8	&	52.2	&	47.8	&	43.4	&	59.6	\\
professional psychology	&	37.7	&	43.0	&	50.2	&	45.1	&	51.0	&	60.6	&	59.5	&	62.4	&	74.0	\\
public relations	&	40.0	&	46.4	&	44.5	&	51.8	&	54.5	&	66.4	&	63.6	&	66.4	&	68.2	\\
security studies	&	35.1	&	33.5	&	38.8	&	44.1	&	39.6	&	57.6	&	60.8	&	61.6	&	72.2	\\
sociology	&	45.3	&	51.2	&	51.2	&	52.7	&	60.2	&	69.2	&	74.1	&	78.6	&	85.1	\\
us foreign policy	&	58.0	&	70.0	&	73.0	&	63.0	&	63.0	&	74.0	&	80.0	&	80.0	&	83.0	\\
virology	&	34.3	&	34.3	&	32.5	&	38.0	&	42.8	&	45.2	&	47.6	&	49.4	&	53.0	\\
world religions	&	65.5	&	69.0	&	71.9	&	70.2	&	82.5	&	80.1	&	83.6	&	83.6	&	87.1	\\
    \bottomrule
    \end{tabular}}
            \caption{MMLU Test set scores for \Atlas{} for each model size and each of the 57 domains.}
    \label{tab:full_mmlu}
\end{table*}

\begin{table*}
    \centering
            \scalebox{0.79}{

    \begin{tabular}{r ccc ccc ccc}
    \toprule
    & \multicolumn{3}{c}{5-shot} & \multicolumn{3}{c}{5-shot (multi-task)} & \multicolumn{3}{c}{Full / Transfer} \\
    \cmidrule(lr){2-4} \cmidrule(lr){5-7} \cmidrule(lr){8-10}
& 770M &  3B  & 11B  & 770M &  3B  & 11B & 770M &  3B  & 11B \\
All	&	29.2	&	35.7	&	36.1	&	26.5	&	40.0	&	43.5	&	42.4	&	50.4	&	54.0	\\
\midrule
Humanities	&	30.5	&	35.4	&	35.5	&	27.3	&	38.5	&	41.6	&	41.0	&	48.6	&	51.3	\\
Social Sciences	&	29.7	&	38.0	&	39.4	&	24.8	&	43.8	&	48.9	&	48.6	&	57.8	&	64.7	\\
STEM	&	29.0	&	31.4	&	30.8	&	26.5	&	32.8	&	35.8	&	33.4	&	40.6	&	41.7	\\
Other	&	26.7	&	37.7	&	38.6	&	27.0	&	45.0	&	48.5	&	46.8	&	55.2	&	59.1	\\
\midrule
abstract algebra	&	26.0	&	23.0	&	21.0	&	29.0	&	30.0	&	26.0	&	23.0	&	29.0	&	26.0	\\
anatomy	&	21.5	&	40.0	&	40.7	&	27.4	&	39.3	&	45.9	&	35.6	&	43.7	&	42.2	\\
astronomy	&	37.5	&	38.8	&	37.5	&	27.6	&	39.5	&	41.4	&	36.2	&	50.7	&	55.3	\\
business ethics	&	29.0	&	54.0	&	42.0	&	26.0	&	47.0	&	55.0	&	53.0	&	64.0	&	60.0	\\
clinical knowledge	&	32.5	&	33.6	&	40.0	&	28.7	&	44.2	&	47.9	&	45.3	&	52.8	&	57.7	\\
college biology	&	29.9	&	34.7	&	34.0	&	29.9	&	34.7	&	40.3	&	38.2	&	46.5	&	52.1	\\
college chemistry	&	37.0	&	22.0	&	32.0	&	20.0	&	35.0	&	33.0	&	36.0	&	34.0	&	36.0	\\
college computer science	&	28.0	&	35.0	&	34.0	&	28.0	&	27.0	&	36.0	&	31.0	&	44.0	&	35.0	\\
college mathematics	&	31.0	&	29.0	&	27.0	&	22.0	&	34.0	&	27.0	&	30.0	&	33.0	&	32.0	\\
college medicine	&	24.3	&	34.7	&	34.1	&	27.2	&	40.5	&	40.5	&	35.8	&	41.6	&	48.6	\\
college physics	&	33.3	&	23.5	&	23.5	&	22.5	&	19.6	&	26.5	&	22.5	&	32.4	&	24.5	\\
computer security	&	36.0	&	42.0	&	46.0	&	31.0	&	49.0	&	52.0	&	50.0	&	65.0	&	61.0	\\
conceptual physics	&	26.4	&	35.7	&	30.2	&	23.4	&	30.6	&	32.8	&	34.5	&	37.4	&	43.8	\\
econometrics	&	26.3	&	21.9	&	28.9	&	17.5	&	19.3	&	24.6	&	29.8	&	25.4	&	29.8	\\
electrical engineering	&	31.0	&	33.1	&	31.7	&	31.0	&	31.0	&	36.6	&	41.4	&	47.6	&	51.7	\\
elementary mathematics	&	26.2	&	27.5	&	28.0	&	27.0	&	31.2	&	33.3	&	25.9	&	31.2	&	35.5	\\
formal logic	&	34.1	&	34.1	&	31.7	&	15.1	&	34.9	&	31.0	&	31.7	&	38.1	&	42.1	\\
global facts	&	32.0	&	30.0	&	25.0	&	34.0	&	34.0	&	27.0	&	28.0	&	34.0	&	30.0	\\
high school biology	&	22.6	&	31.9	&	29.7	&	27.1	&	41.6	&	50.0	&	43.5	&	57.7	&	60.6	\\
high school chemistry	&	27.1	&	26.6	&	27.6	&	28.6	&	31.5	&	29.1	&	30.5	&	36.5	&	38.9	\\
high school computer science	&	26.0	&	32.0	&	25.0	&	33.0	&	37.0	&	45.0	&	45.0	&	55.0	&	48.0	\\
high school european history	&	34.5	&	43.0	&	42.4	&	24.2	&	60.0	&	59.4	&	58.2	&	69.1	&	76.4	\\
high school geography	&	31.3	&	40.4	&	36.9	&	24.7	&	45.5	&	50.5	&	56.1	&	66.7	&	74.2	\\
high school gov. and pol.	&	28.0	&	49.2	&	51.3	&	19.2	&	56.0	&	59.6	&	55.4	&	70.5	&	75.6	\\
high school macroeconomics	&	25.6	&	37.7	&	32.1	&	26.7	&	42.3	&	43.6	&	41.0	&	51.5	&	56.4	\\
high school mathematics	&	35.9	&	35.2	&	35.9	&	28.1	&	26.7	&	31.1	&	27.8	&	36.7	&	31.9	\\
high school microeconomics	&	27.3	&	29.8	&	36.1	&	20.6	&	35.7	&	42.9	&	42.9	&	50.8	&	60.5	\\
high school physics	&	21.9	&	25.2	&	22.5	&	24.5	&	28.5	&	29.1	&	27.8	&	31.1	&	27.8	\\
high school psychology	&	26.1	&	46.4	&	51.0	&	24.8	&	54.3	&	60.2	&	56.3	&	67.3	&	76.1	\\
high school statistics	&	27.8	&	33.3	&	33.3	&	17.6	&	30.6	&	33.8	&	32.9	&	33.3	&	37.0	\\
high school us history	&	30.4	&	39.7	&	45.6	&	27.5	&	46.1	&	58.3	&	51.0	&	63.2	&	72.5	\\
high school world history	&	42.6	&	50.6	&	41.8	&	29.1	&	54.0	&	64.6	&	66.7	&	72.2	&	73.8	\\
human aging	&	28.3	&	37.2	&	29.6	&	26.0	&	45.3	&	46.2	&	46.6	&	57.0	&	62.8	\\
human sexuality	&	29.8	&	34.4	&	41.2	&	25.2	&	42.0	&	44.3	&	51.1	&	58.0	&	59.5	\\
international law	&	57.9	&	57.9	&	41.3	&	44.6	&	57.9	&	58.7	&	62.8	&	71.9	&	71.1	\\
jurisprudence	&	30.6	&	33.3	&	34.3	&	32.4	&	49.1	&	52.8	&	55.6	&	67.6	&	74.1	\\
logical fallacies	&	40.5	&	55.8	&	46.6	&	25.8	&	51.5	&	62.0	&	43.6	&	69.3	&	71.2	\\
machine learning	&	33.0	&	34.8	&	36.6	&	29.5	&	35.7	&	37.5	&	32.1	&	37.5	&	42.9	\\
management	&	21.4	&	29.1	&	40.8	&	24.3	&	47.6	&	50.5	&	60.2	&	69.9	&	70.9	\\
marketing	&	38.9	&	58.5	&	60.7	&	31.2	&	67.9	&	75.6	&	69.2	&	79.9	&	85.9	\\
medical genetics	&	26.0	&	36.0	&	36.0	&	29.0	&	43.0	&	44.0	&	40.0	&	54.0	&	50.0	\\
miscellaneous	&	24.5	&	45.2	&	46.4	&	27.1	&	52.2	&	58.2	&	51.3	&	64.6	&	72.7	\\
moral disputes	&	32.4	&	37.3	&	38.7	&	28.6	&	43.4	&	43.4	&	49.7	&	64.7	&	64.7	\\
moral scenarios	&	24.7	&	24.7	&	24.7	&	23.0	&	23.9	&	24.7	&	23.8	&	24.0	&	23.8	\\
nutrition	&	30.1	&	33.0	&	34.6	&	25.8	&	42.5	&	44.1	&	50.3	&	55.6	&	61.1	\\
philosophy	&	28.6	&	32.5	&	37.3	&	31.2	&	38.9	&	45.0	&	44.1	&	56.6	&	59.2	\\
prehistory	&	33.6	&	37.0	&	41.4	&	27.5	&	39.8	&	50.6	&	41.0	&	51.5	&	57.7	\\
professional accounting	&	21.3	&	28.0	&	30.5	&	25.9	&	35.5	&	34.0	&	37.2	&	41.5	&	42.2	\\
professional law	&	28.2	&	33.4	&	34.0	&	27.6	&	35.4	&	35.5	&	38.3	&	43.0	&	45.6	\\
professional medicine	&	19.5	&	26.5	&	24.3	&	20.2	&	32.0	&	37.9	&	38.6	&	40.8	&	46.0	\\
professional psychology	&	27.8	&	32.8	&	32.8	&	26.6	&	39.5	&	43.6	&	38.4	&	48.0	&	58.3	\\
public relations	&	22.7	&	43.6	&	40.0	&	21.8	&	47.3	&	56.4	&	50.0	&	55.5	&	60.0	\\
security studies	&	37.6	&	26.1	&	31.0	&	20.4	&	34.7	&	44.1	&	56.3	&	61.6	&	66.9	\\
sociology	&	43.3	&	41.8	&	38.8	&	30.8	&	45.8	&	52.7	&	60.2	&	66.7	&	72.1	\\
us foreign policy	&	49.0	&	57.0	&	66.0	&	38.0	&	56.0	&	61.0	&	59.0	&	75.0	&	76.0	\\
virology	&	29.5	&	26.5	&	34.3	&	30.1	&	36.1	&	39.8	&	44.0	&	46.4	&	41.6	\\
world religions	&	24.0	&	40.9	&	47.4	&	32.7	&	49.1	&	57.3	&	48.0	&	63.7	&	70.2	\\
    \bottomrule
    \end{tabular}}
    \caption{MMLU Test scores for the T5 closed book baseline for each model size and each of the 57 domains.}
    \label{tab:full_mmlu_cb}
\end{table*}

\begin{table*}
    \centering
            \scalebox{0.79}{

    \begin{tabular}{r cccc}
    \toprule
Domain    & \hspace*{5mm} zero-shot  \hspace*{5mm}& \hspace*{5mm} 5-shot \hspace*{5mm} & 5-shot (multi-task)&  Full / Transfer \\
    \midrule
All	&	47.1	&	47.9	&	56.6	&	66.0	\\
\midrule
Humanities	&	43.6	&	46.1	&	50.1	&	61.1	\\
Social Sciences	&	54.1	&	54.6	&	66.4	&	77.2	\\
STEM	&	38.0	&	38.8	&	46.4	&	53.2	\\
Other	&	53.9	&	52.8	&	66.2	&	74.4	\\
\midrule
abstract algebra	&	22.0	&	26.0	&	31.0	&	31.0	\\
anatomy	&	48.9	&	47.4	&	62.2	&	70.4	\\
astronomy	&	61.8	&	62.5	&	68.4	&	81.6	\\
business ethics	&	60.0	&	57.0	&	62.0	&	70.0	\\
clinical knowledge	&	50.6	&	49.4	&	66.4	&	72.8	\\
college biology	&	51.4	&	53.5	&	61.1	&	77.8	\\
college chemistry	&	36.0	&	39.0	&	39.0	&	45.0	\\
college computer science	&	32.0	&	32.0	&	33.0	&	49.0	\\
college mathematics	&	30.0	&	35.0	&	35.0	&	34.0	\\
college medicine	&	44.5	&	41.0	&	52.6	&	67.6	\\
college physics	&	24.5	&	26.5	&	37.3	&	42.2	\\
computer security	&	59.0	&	59.0	&	68.0	&	76.0	\\
conceptual physics	&	37.0	&	41.3	&	57.0	&	60.0	\\
econometrics	&	20.2	&	20.2	&	36.8	&	37.7	\\
electrical engineering	&	37.9	&	40.0	&	50.3	&	65.5	\\
elementary mathematics	&	31.2	&	28.0	&	30.7	&	36.5	\\
formal logic	&	27.8	&	27.0	&	32.5	&	35.7	\\
global facts	&	41.0	&	43.0	&	51.0	&	53.0	\\
high school biology	&	53.2	&	56.5	&	68.7	&	83.2	\\
high school chemistry	&	41.9	&	41.4	&	49.3	&	51.2	\\
high school computer science	&	40.0	&	36.0	&	46.0	&	60.0	\\
high school european history	&	56.4	&	58.8	&	68.5	&	80.6	\\
high school geography	&	57.1	&	59.6	&	71.2	&	81.3	\\
high school gov. and pol.	&	67.9	&	67.9	&	77.2	&	90.2	\\
high school macroeconomics	&	46.9	&	48.5	&	57.9	&	65.9	\\
high school mathematics	&	28.1	&	28.9	&	34.1	&	31.5	\\
high school microeconomics	&	51.7	&	51.7	&	68.9	&	82.4	\\
high school physics	&	26.5	&	25.8	&	32.5	&	41.1	\\
high school psychology	&	66.2	&	65.5	&	78.9	&	86.8	\\
high school statistics	&	31.5	&	30.1	&	43.1	&	45.8	\\
high school us history	&	57.8	&	54.9	&	64.7	&	77.5	\\
high school world history	&	59.1	&	62.9	&	65.4	&	79.3	\\
human aging	&	48.4	&	50.7	&	60.5	&	70.4	\\
human sexuality	&	55.7	&	54.2	&	61.8	&	84.0	\\
international law	&	66.1	&	72.7	&	71.9	&	84.3	\\
jurisprudence	&	61.1	&	64.8	&	72.2	&	81.5	\\
logical fallacies	&	54.6	&	57.7	&	71.2	&	77.9	\\
machine learning	&	37.5	&	39.3	&	43.8	&	44.6	\\
management	&	56.3	&	56.3	&	79.6	&	89.3	\\
marketing	&	72.2	&	73.1	&	84.6	&	91.9	\\
medical genetics	&	55.0	&	58.0	&	71.0	&	81.0	\\
miscellaneous	&	69.7	&	67.8	&	83.8	&	90.4	\\
moral disputes	&	45.1	&	46.8	&	60.1	&	72.3	\\
moral scenarios	&	24.5	&	30.3	&	25.8	&	38.5	\\
nutrition	&	56.5	&	53.9	&	67.0	&	77.1	\\
philosophy	&	56.3	&	57.6	&	70.7	&	77.2	\\
prehistory	&	59.3	&	60.5	&	71.6	&	78.7	\\
professional accounting	&	35.1	&	33.0	&	42.2	&	50.7	\\
professional law	&	36.3	&	38.4	&	39.4	&	51.7	\\
professional medicine	&	35.7	&	33.1	&	52.2	&	60.7	\\
professional psychology	&	47.7	&	49.3	&	60.9	&	74.0	\\
public relations	&	54.5	&	53.6	&	68.2	&	68.2	\\
security studies	&	47.3	&	45.7	&	59.2	&	73.9	\\
sociology	&	62.2	&	62.7	&	71.6	&	84.6	\\
us foreign policy	&	64.0	&	68.0	&	73.0	&	83.0	\\
virology	&	39.8	&	40.4	&	44.6	&	51.8	\\
world religions	&	77.2	&	74.9	&	80.7	&	87.1	\\
    \bottomrule
    \end{tabular}}
    \caption{MMLU Test set scores for the de-biased \Atlas{}-XXL using cyclic permutations for each of the 57 domains for zero-shot, 5 shot, 5-shot-multitask and the transfer setting.}
    \label{tab:full_mmlu_debias}
\end{table*}

\subsection{Question answering}
\label{app:qa}
\subsubsection{Training Details}
For question answering, similarly to the MMLU experiments, we format the input using the following template:

\texttt{question: \{question text\} answer: [MASK\_0]}

and train the model to generate the mask token followed by the answer:

\texttt{[MASK\_0] \{answer\}}.

We generate answers using greedy decoding.
For both training and testing, we retrieve 40 passages, and truncate the result of the concatenation between the query and the passages to 384 tokens.

For few-shot fine-tuning we train \Atlas{} for 30 steps using 64 random samples from the train sets.
The retriever is trained using query-side fine-tuning.
We select the model after 30 training steps.
We use AdamW with a batch size of 32 and a learning rate of $4\times 10^{-5}$ with linear decay and 5 iterations of warmup for both the language model and the retriever.

For the fine-tuning on the full datasets, we train the model for 5k gradient steps and refresh the index every 500 steps for the first 1,000 training steps and every 2k training steps afterwards.
We use AdamW with a batch size of 64 and a learning rate of $4\times 10^{-5}$ with linear decay and 5 iterations of warmup for both the language model and the retriever.
We evaluate models every 500 steps and select the best one on the validation set based on the exact match score.

\subsubsection{Impact of scaling}
In Table~\ref{tab:qa_size}, we report performance on NaturalQuestions and TriviaQA as a function of the number of parameters in the reader module.
Both for few-shot learning and full fine-tuning we observe strong improvements by scaling the size of the reader module.
However we can notice sign of saturation when finetuning on full datasets, with limited gains when scaling from 3B to 11B parameters (+0.6\% on NaturalQuestions, +0.5\% on TriviaQA).
While performance improves substantially when scaling from 3B to 11B parameters with 64 training samples, with +3.7\% and +1.2\% improvement on NaturalQuestions and TriviaQA respectively.
For these experiments we use a setup similar to the one use in Table~\ref{table:nq_sota}, except that we use an index composed of the December 2018 Wikipedia dump processed as described in section~\ref{sec:training}.

\begin{table*}

    \centering
    \begin{tabular}{lcccc}
    \toprule
    Number of parameters & 220M & 770M & 3B & 11B \\
    \midrule
    NaturalQuestions 64-shot & 27.0 & 35.4 & 41.3 & 45.1 \\
    NaturalQuestions full & 54.1 & 60.8 & 63.4 & 64.0 \\ 
    \midrule
    TriviaQA 64-shot & 55.3 & 65.0 & 70.2 & 71.4 \\
    TriviaQA full & 71.8 & 74.9 & 77.5 & 78.0 \\
    \bottomrule
    \end{tabular}
    \caption{
    \textbf{Impact of model size on question answering datasets.} We report exact match performance on the test sets of NaturalQuestions and TriviaQA filtered depending on the number of parameters in the reader module. For these experiments the index contains the December 2018 Wikipedia dump.}
    \label{tab:qa_size}
\end{table*}

\subsection{KILT}
\label{app:kilt}
For the results on KILT reported in Table~\ref{tab:kilt_sota} we fine-tune \Atlas{} individually on each dataset.
We format the input using a template similar to the one used for question answering:

\texttt{question: \{query text\} answer: [MASK\_0]}

and train the model to generate the mask token followed by the expected output:

\texttt{[MASK\_0] \{output\}}.

We retrieve 20 passages and generate answer using greedy decoding.
In KILT, FEVER is a two-way classification task of claims.
We lexicalize the ``SUPPORTS'' (resp. `REFUTES'') label into ``true'' (respectively ``false'').

For few-shot fine-tuning we train \Atlas{} for 30 steps using 64 random samples from the train sets.
The retriever is trained using query-side fine-tuning.
We evaluate models every 5 steps and select the best one on the development set based on the reported metric.
We use AdamW with a batch size of 32 and a learning rate of $4\times 10^{-5}$ with linear decay and 5 iterations of warmup for both the language model and the retriever.

For the fine-tuning on the full datasets, the model is trained for 5k gradient steps.
We evaluate models every 500 steps and select the best one on the development set based on the reported metric.
The index is refreshed every 500 step for the first 1000 iterations, and every 2k steps afterwards.
We use AdamW with a batch size of 64 and a learning rate of $4\times 10^{-5}$ with linear decay and 500 iterations of warmup for both the language model and the retriever.

We report results on the development sets in Table~\ref{tab:kilt_dev}.

\begin{table*}
    \centering
    \setlength{\tabcolsep}{4pt}
    \begin{tabular}{lcccccccccc}
    \toprule
    \multirow{2}{*}{Model}         & {AIDA} & {FEV} & {T-REx} & {zsRE} & {NQ} & {HoPo} & {TQA} & {WoW}  \\
    & {\textsc{acc}} & {\textsc{acc}} & {\textsc{acc}} & {\textsc{acc}} & {\textsc{em}}   & {\textsc{em}}   & {\textsc{em}} & {\textsc{f1}}   \\ \midrule
    \Atlas{} 64-shot  & 69.0 & 88.1 & 58.5 & 60.2 & 44.2 & 34.1 & 77.1 & 15.4 \\
    \Atlas{} full dataset & 92.7 & 94.4 & 84.8 & 80.9 & 63.4 & 51.4 & 84.4 & 21.0 \\
    \bottomrule
    \end{tabular}
    \caption{\textbf{Downstream results on KILT dev sets.}}
    \label{tab:kilt_dev}
\end{table*}


\end{document}